\documentclass{article} % For LaTeX2e
\usepackage[preprint]{colm2025_conference}
\usepackage{pifont}

\usepackage[T1]{fontenc}    % fix “OMS/zi4/m/n undefined” by enabling T1 encoding
\usepackage{textcomp}       % supplemental symbols
\usepackage{lmodern}  
\usepackage{cfr-lm}
\usepackage{inconsolata}    % actually load the zi4 (Inconsolata) tt font

\usepackage{graphicx}
\newcommand{\Bench}{\textsc{SimCopilot}}
\newcommand{\BenchJ}{\textsc{SimCopilotJ}}
\newcommand{\BenchP}{\textsc{SimCopilotP}}

\usepackage{microtype}
\usepackage{hyperref}
\usepackage{url}
\usepackage{bm}
\usepackage{caption}
\usepackage{booktabs}
\usepackage{wrapfig}
\usepackage{multirow}
\usepackage{listings}
\usepackage{enumitem}
\usepackage{algorithm}
\usepackage{algorithmic}
\usepackage{subcaption}
\usepackage{multicol}
\usepackage{listings}
\usepackage{xcolor}
\usepackage{lineno}
\lstdefinestyle{javaStyle}{
    language=Java,
    basicstyle=\tiny\ttfamily,
    keywordstyle=\color{blue}\bfseries,
    stringstyle=\color{red},
    commentstyle=\color{gray},
    numbers=none,
    backgroundcolor=\color{white},
    showstringspaces=false,
    tabsize=4,
    morekeywords={var,let,const} % optional: add extra keywords if needed
}
\definecolor{darkblue}{rgb}{0, 0, 0.5}
\hypersetup{colorlinks=true, citecolor=darkblue, linkcolor=darkblue, urlcolor=darkblue}

\newcommand{\cmdfontsize}{\fontsize{5pt}{5pt}\selectfont}  % Custom font size (< \tiny < \scriptsize < \footnotesize < \small < \normal)
\newcommand{\cmdfontnumsize}{\fontsize{3.5pt}{3.5pt}\selectfont} 
\definecolor{lightyellow}{rgb}{0.88, 0.88, 0} % A light shade of red

\lstdefinestyle{javaStyle}{
  language=Java,
  basicstyle=\cmdfontsize\ttfamily,
  keywordstyle=\color{blue}\bfseries,
  commentstyle=\color{green!70!black}\itshape,
  stringstyle=\color{orange},
  numbers=left,
  numberstyle=\cmdfontnumsize\ttfamily\color{gray},
  stepnumber=1,
  numbersep=5pt,
  tabsize=2,
  showspaces=false,
  showstringspaces=false,
  breaklines=true,
  breakatwhitespace=true,
  frame=single,
  backgroundcolor=\color{gray!10},  % Shade of gray that is 10% black and 90% white
  captionpos=b,
  morekeywords={your,additional,keywords,here}
}

\lstdefinestyle{pythonStyle}{
    language=Python,
    basicstyle=\cmdfontsize\ttfamily,
    keywordstyle=\color{blue}\bfseries,
    commentstyle=\color{green!70!black}\itshape,
    stringstyle=\color{orange},
    numberstyle=\cmdfontnumsize\ttfamily\color{gray},
    backgroundcolor=\color{lightgray!20},
    numbers=left,
    stepnumber=1,
    numbersep=5pt,
    showspaces=false,
    showstringspaces=false,
    showtabs=false,
    frame=single,
    tabsize=4,
    captionpos=b,
    breaklines=true,
    breakatwhitespace=true,
    morekeywords={your,additional,keywords,here}
}

\newlist{rqitemize}{enumerate}{1}
\setlist[rqitemize,1]{label=RQ\arabic*,left=0.5pt}

\title{\Bench{}: Evaluating Large Language Models \\ for Copilot-Style Code Generation}

\author{Mingchao Jiang$^1$, Abhinav Jain$^1$, Sophia Zorek$^2$ \& Chris Jermaine$^1$ 
 \thanks{Code: \href{https://github.com/mj33rice/Sim-CoPilot}{github.com/mj33rice/Sim-CoPilot} • Dataset: \href{https://huggingface.co/datasets/mj33/SimCoPilot}{huggingface.co/datasets/mj33/SimCoPilot}}
 \\[2mm]
$^1$Department of Computer Science 
$^2$Department of Statistics \\[1mm]
Rice University\\
Houston, TX 77005, USA \\
\texttt{\{mj33,aj70,saz2,cmj4\}@rice.edu} \\
}

\begin{document}

\ifcolmsubmission
\linenumbers
\fi

\maketitle

\begin{abstract}
We introduce \Bench{}, a benchmark that simulates the role of large language models (LLMs) as interactive, ``copilot''-style coding assistants. Targeting both \emph{completion} (finishing incomplete methods or code blocks) and \emph{infill} tasks (filling missing segments within existing code), \Bench{} provides a comprehensive framework for evaluating LLM coding capabilities. The benchmark comprises dedicated sub-benchmarks for Java (\BenchJ{}) and Python (\BenchP{}), covering diverse codebases varying in size and complexity. Our key contributions include: (a) establishing a realistic, detailed evaluation environment to assess LLM utility in practical coding scenarios, and (b) providing fine-grained analyses that address critical factors frequently overlooked by existing benchmarks, such as task-specific performance nuances, contextual understanding across code segments, and sensitivity to variable scope. Evaluations conducted across domains—including algorithms, databases, computer vision, and neural networks—offer insights into model strengths and highlight persistent challenges in maintaining logical consistency within complex dependency structures. Beyond benchmarking, our study sheds light on the current limitations of LLM-driven code generation and underscores the ongoing transition of LLMs from merely syntax-aware generators toward reliable, intelligent software development partners. 

\end{abstract}
\section{Introduction}

Currently, the most widely-used benchmarks for checking the ability of AI models to perform program synthesis (``AI-for-code'') consist of a detailed English description of a concise, self-contained code to synthesize, as well as a few test cases to test the correctness of the synthesized code \citep{austin2021program, hendrycks2021measuring, chen2021evaluating,iyer-etal-2018-mapping}.  While such benchmarks are useful, they match one particularly narrow use case, where the goal is to synthesize a relatively short, complete, standalone program.

Arguably, the most impactful and widely-used application of AI-for-code to date has been through tools such as GitHub's Copilot \citep{copilot}. Copilot is designed to be used interactively, where the AI is repeatedly given code completion tasks: given a partially-completed code such as a commented method header with an empty body, an \texttt{if}-statement with an empty \texttt{else} block, or an empty \texttt{for} loop body, can an AI tool correctly complete the next few lines of code?  

Rather than writing code from scratch, the goal is to interactively help a programmer to write code, reducing the programmer's burden and increasing productivity.  
In practice, this is often done in the context of a large software project, where the AI needs to figure out how to call internal APIs that are specific to the project correctly or to reference classes or variables that were declared remotely from the site location of the code completion task.

In this paper, we present an AI-for-code benchmark called \Bench{}, which is specifically designed to simulate a ``Copilot''-style, interactive environment. All of the code completion tasks were chosen so that it would be impossible for an existing AI system to have trained on the code.
\Bench{} consists of two sub-benchmarks, \BenchJ{} (Java) and \BenchP{} (Python).  \BenchJ{} and \BenchP{} are separately designed to test how Java (and related languages) and Python (and related languages) are likely to be used in practice. 

Crucially, \Bench{} is designed to test two different sub-cases in interactive code completion: \emph{completion} and \emph{infill}.  Completion simulates the case where a programmer is using an interactive AI programming tool to write a method/function from start to finish. Here, a method or function is either empty (consisting of only a commented header), or it may have some code starting from the beginning of the method/function, but the code is not complete. The goal is to follow instructions and write a certain portion of the missing code (such as finishing the current \texttt{else} block), following the last-provided line.  However, not all code is written linearly, from start to finish, and \emph{infill} is designed to simulate this case.  In infill, there is a ``blank'' in a method, function, or logic block, and the goal is to provide the missing code.

Java is a structured, relatively high-performance, compiled language generally used to develop larger software projects, whereas Python is a scripting language often used to solve smaller programming problems.  Hence, \BenchJ{} consists of $286$ code completion and $283$ code infilling tasks over eight separate modules (groups of related classes) in a medium-sized, well-documented Java project ($7,728$ lines in all).  \BenchP{} consists of $212$ code completion and $382$ code infilling tasks over $7$ separate Python programs for reasonably standard tasks in computer vision, databases, optimization, etc., of intermediate length (consisting of an average of $431$ lines each).  An AI programming assistant is considered to have ``passed'' one of the code completion tasks if the AI-written code, together with the rest of the code in the program/project, is able to pass an extensive set of test cases.

In this paper, we carefully describe the \Bench{} benchmark and evaluate a number of different AI-for-code models.  The results presented show how \Bench{} can have a much more detailed and meaningful view of the differences between various AI-for-code models.  For example, popular benchmarks such as HumanEval \citep{chen2021evaluating} and MBPP \citep{austin2021program} show relatively modest differences between models. Claude $3$ Opus \citep{claude-all-models}
scores $84.9\%$ on HumanEval, whereas Llama $3.3$ $70$B \citep{llama3-3} scores $88.4\%$. Based on HumanEval alone, one might conclude that Llama $3.3$ $70$B outperforms Claude $3$ Opus. However, \Bench{} reveals a different picture across more realistic programming tasks. Claude $3$ Opus achieves $67.4\%$ and $69.2\%$ on Python and Java completion tasks. In contrast, Llama $3.3$ $70$B scores significantly lower performance of $53.7\%$ and $49.3\%$ on Python and Java completion tasks. This dramatic reversal demonstrates how \Bench{} provides a more comprehensive and realistic assessment of model capabilities in practical programming scenarios, where contextual understanding and code integration skills are essential.

\section{Related Work}
\begin{figure*}[t]
     \centering
\includegraphics[width=0.9\textwidth]{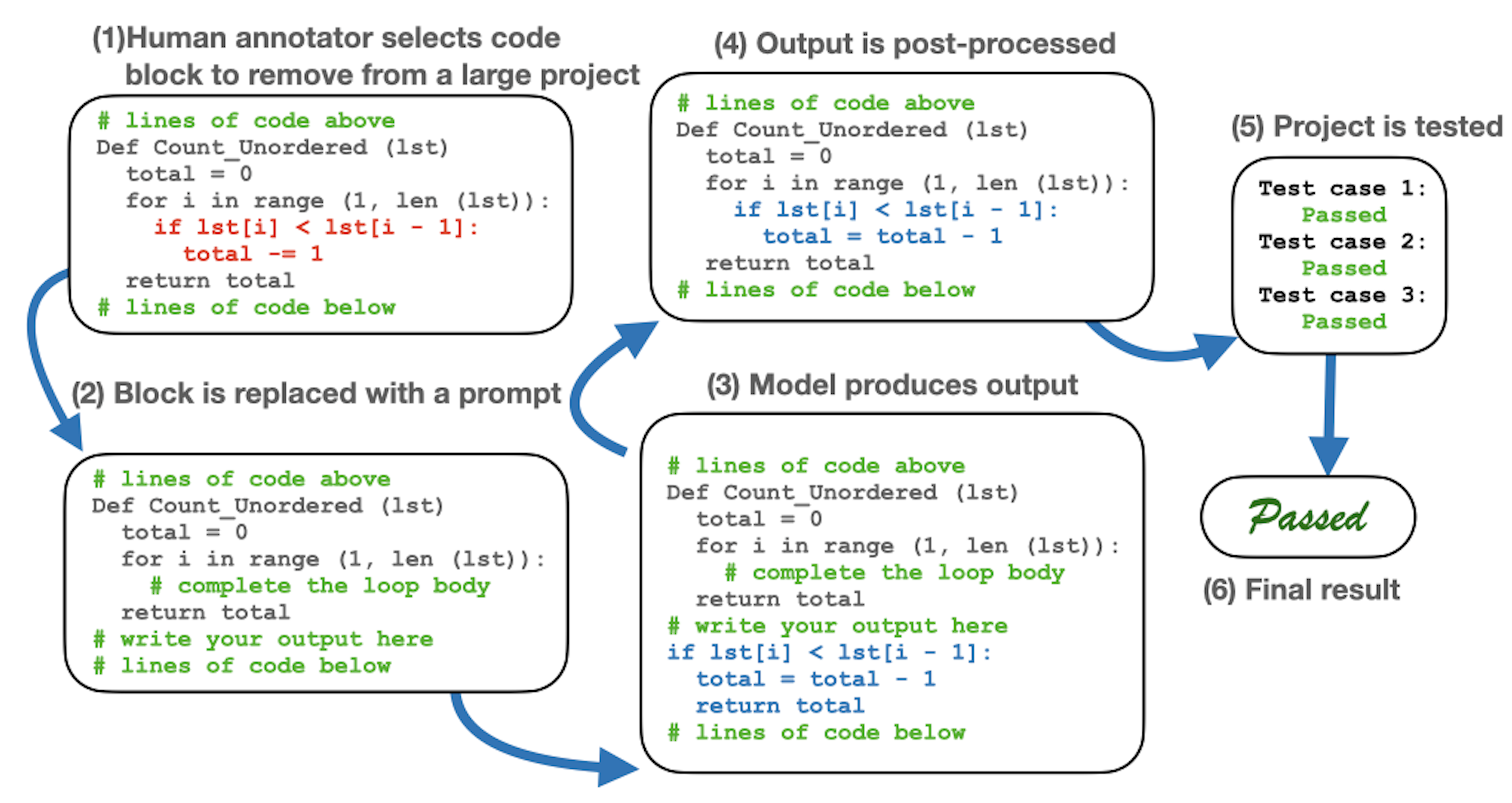}
     \caption{Workflow for each of the $1,163$ programming tasks in \Bench{}. }
     \label{fig:frame}
\end{figure*}
Recent code generation benchmarks have evolved from generating concise, standalone programs from detailed specifications, as seen in HumanEval \citep{chen2021evaluating}, MBPP \citep{austin2021program}, and APPS \citep{hendrycks2021measuring}, to more context-aware evaluations. CrossCodeEval \citep{ding2024crosscodeeval} and RepoBench \citep{liu2023repobench} assess code completion within larger repositories, focusing on cross-file references. ClassEval \citep{du2023classeval}, CoderEval \citep{yu2024codereval}, and EvoCodeBench \citep{li2024evocodebench} emphasize test-based validation yet evaluate AI models using detailed, manually annotated problem descriptions to prompt code generation. SWE-bench Verified \citep{openai2024swebenchverified} tasks models with resolving real-world GitHub issues and validates their patches against regression tests. LiveCodeBench \citep{jain2024livecodebench} offers dynamic, contamination-free challenges from online contests, assessing capabilities like iterative self-repair and code execution. ProBench \citep{yang2025probench} focuses on competitive programming tasks, analyzing reasoning depth and error patterns. DynaCode \citep{hu2025dynacode} introduces complexity-aware evaluations with nested code tasks, while BigO(Bench) \citep{chambon2025bigo} evaluates adherence to specified time and memory efficiency constraints.

Despite these comprehensive benchmarks, none directly simulate the interactive ``AI pair programmer'' workflow. \Bench{} addresses this gap by focusing on left-to-right code completion and mid-code infilling tasks, reflecting how developers use tools like Copilot to incrementally add and refine code. The benchmark provides a realistic, detailed evaluation environment for LLMs acting as coding assistants, enabling fine-grained analyses of underexplored challenges such as scope sensitivity, task-specific performance variations, and inter-segment contextual dependencies. 
These contributions distinguish \Bench{} among code generation benchmarks by targeting practical, context-dependent, interactive coding assistance scenarios.

\section{\Bench{} Benchmark Design}
\label{sec:bench_design}

\Bench{} consists of $569$ Copilot-style Java programming problems, and $594$ Python problems, as well as a comprehensive suite of software tools for the ancillary tasks necessary to run the benchmark, such as code post-processing and data analysis.  The general workflow for creating and executing each programming problem is shown in Figure \ref{fig:frame}. A human annotator chooses a section of code to omit (the benchmark comes with $1,163$ sections of code marked for omission).  When it is time to test an AI-for-code model, the \Bench{} software removes the annotated code and replaces it with a prompt. The model is asked to write the missing code; the model's output is then post-processed and inserted back into the original code, which is then tested.  If the code passes all test cases, the model passes (See as Fig.\ref{fig:python_processing} and \ref{fig:java_processing} in appendix).

\begin{table}[!t]
\centering

  \begin{tabular}{lcccc}
  \multirow{2}{*}{} & \multicolumn{2}{c}{Python} & \multicolumn{2}{c}{Java} \\
  \cmidrule(lr){2-3} \cmidrule(lr){4-5}
  & Infill & Compl. & Infill & Compl. \\
  \midrule
  Total Count       & $382$  & $212$ & $283$  & $286$  \\
  \midrule
  Local Variable    & $371$  & $206$ & $232$  & $218$  \\
  Global Variable   & -    & -   & $157$  & $184$  \\
  Function          & $37$   & $9$   & $110$  & $88$   \\
  Class             & $15$   & $19$  & $54$   & $32$   \\
  Library           & $202$  & $123$ & $6$    & $10$   \\
  If-Else Cond      & $19$   & $14$  & $56$   & $33$   \\
  If-Else Body      & $61$   & $73$  & $79$   & $120$  \\
  Loop Body         & $72$   & $49$  & $102$  & $102$  \\
  \midrule
  Avg Num Lines     & \multicolumn{2}{c}{$431$} & \multicolumn{2}{c}{$966$} \\
  \bottomrule
  \end{tabular}
\captionsetup{hypcap=false}
\caption{Task categories and frequencies.}
\captionsetup{hypcap=true}
\label{tab:task-freq}
\end{table}

\subsection{Annotation}

Each of the $1,163$ programming tasks was created from eight Java repositories and seven Python repositories, totaling nearly $11,000$ lines of code. We employed a systematic two-stage pipeline to identify semantically meaningful code blocks for generation tasks:

\begin{itemize}[leftmargin=*,nosep]
    \item \textbf{AST-based parsing:} We construct an abstract syntax tree (AST) for each source file to locate constructs such as classes, functions, loops, and conditionals.
    \item \textbf{DFS-based block extraction:} We apply a depth-first traversal to extract complete, self-contained blocks, ensuring syntactic integrity and well-formed prompts.
\end{itemize}

To enforce consistency, block selection is driven by deterministic structural rules. 
Specifically, we \ding{192} exclude trivial boilerplates, such as 1-line accessors or stubs, \ding{193} filter for non-trivial logic with control flow or function calls, and \ding{194} ensure broad task coverage, including control structures, function or library calls, and object references.

% Specifically, we \textcircled{1} exclude trivial boilerplates, such as 1-line accessors or stubs, \textcircled{2} filter for non-trivial logic with control flow or function calls, and \textcircled{3} ensure broad task coverage, including control structures, function or library calls, and object references.

Our team, consisting of PhD-level programmers, went through these codes, generating both infill and completion tasks.  To create an infill task, the annotator picks a meaningful starting point for the AI-for-code model to begin writing code (at the beginning of the boolean \texttt{if} condition, or at the beginning of the body of a \texttt{for} loop, for example see as Fig.\ref{fig:python_processing}) and then marks the rest of that particular code block for deletion, to be re-created by the AI-for-code model.  In the case of an \texttt{if} condition, the entire boolean predicate would be marked for deletion.  In the case of a \texttt{for} loop body, the entire body would be marked.  A completion task is created in much the same way, but the code for the remainder of the method or function is marked for deletion (see as Fig.\ref{fig:java_processing}). In total, the average number of lines deleted for infill is $2.62$ for Java and $2.52$ for Python, and the average number of lines deleted for completion is $3.31$ for Java and $3.89$ for Python.

One consideration when choosing code blocks for deletion is that we want good coverage of various programming tasks, as \Bench{} does not simply report the overall pass rate but breaks all of the $1,163$ programming tasks into eight overlapping categories and reports results from each.  These categories are as follows: (1) \emph{Local variable}---the programming task requires the use of a previously-defined local variable; (2) \emph{Global variable}---the task requires the use of a global variable; (3) \emph{Function}---requires calling a function defined previously in the repository; (4) \emph{Class}---requires a reference to a class previously defined; (5) \emph{Library}---requires use of an external library; (6) \emph{If-else-cond}---requires generating the boolean condition in an \texttt{if} or \texttt{else if} statement; (7) \emph{If-else-body}---requires generating the body of an \texttt{if},  \texttt{else if}, \texttt{else} statement; (8) \emph{Loop body}---requires generating a loop body.  The number of occurrences of each of these eight different types of tasks is given in Table \ref{tab:task-freq}.

\subsection{Pre-Processing}

To generate an actual programming task, the \Bench{} software locates a block marked for deletion.  All other files in the repository are prepended to the file containing the marked block, the class/function/method containing the marked block is moved to the end of the file, and the marked block is replaced with a prompt.  Then, the AI-for-code model reacts to the prompt and attempts to re-create the missing code.  All of this is done so as to mimic how a code completion task might be prepared by a Copilot-style software deployment.

Our current version of the \Bench{} software contains only one generic pre-processing pipeline.  However, as the \Bench{} software evolves and the benchmark is applied to different AI-for-code models, it is acceptable---even desirable---for the authors or particular AI-for-code models to prepare their own pre-processing pipelines that work best with their own model.  Our reasoning is that in a ``real-life'' deployment where an AI-for-code model is used to write code interactively, there is undoubtedly going to be a highly specialised pre-processing pipeline that works well with the specific model.

Our current prompt emphasizes the following aspects of code generation to the AI-for-code model and was engineered through trial and error: (1) The code should logically continue from the section before the prompt (and connect smoothly to the section after the prompt for infill tasks); (2) Insert only syntactically correct code without any additional comments or text; (3) Ensure all brackets, parentheses, and curly braces are properly paired and closed (and matches indentation for Python); (4) For completion tasks, complete the innermost incomplete method, function, loop body, if-statement, or self-contained code block.

\subsection{Post-Processing}

Post-processing is crucial, as even the best AI-for-code models add irrelevant English or other comments and repeat lines of code, or even whole code blocks, despite being prompted not to. Any such model deployed in a production environment would have a custom-built post-processor to attempt to correct such problems.  The \Bench{} benchmark comes with a post-processor, though different post-processors customized for specific AI-for-code models may be used.

Post-processing begins by using a series of regular expressions to eliminate statements that are clearly not statements in the target programming language.  Post-processing then removes any repeating lines of code, or repeating code blocks.  One very common problem is when the AI-for-code model re-generates code above the prompt, or---in an infill task---it re-generates code following the prompt.  Our post-processor searches for both exact and approximate re-generation of provided code, and then removes any such re-generated code from the response. The next step is to process indentation (python) and bracket (Java) discrepancies. For Python, the post-processor automatically aligns the generated code with the preceding code that the AI-for-code model is attempting to complete. For Java, the post-processor adds or deletes brackets from the generated code in an attempt to make it syntactically correct. Once the generated code has been post-processed, it is inserted as a replacement for the deleted code block, and the resulting code is tested for correctness using a series of test cases.

\subsection{Code Repositories Used}
  
We chose a set of Java and Python repositories to use for the benchmark.  None of the repositories were publicly accessible, to make it highly unlikely that any existing AI-for-code model had been trained on the repository.

\begin{table}
    \centering
    \captionsetup{skip=5pt} % Adjust the spacing between the table and caption
    \centerline{
    \begin{tabular}{lcccccc}
        \toprule
        % \multirow{2}{*}{Model} & \multicolumn{2}{c}{Python} & \multicolumn{2}{c}{Java} & \multirow{2}{*}{Hum} & \multirow{2}{*}{SWE} \\
        \multirow{2}{*}{Model} & \multicolumn{2}{c}{Python} & \multicolumn{2}{c}{Java} & \multirow{2}{*}{Hum-}  \\
\cmidrule(lr){2-3} \cmidrule(lr){4-5}
& Infill & Compl. & Infill & Compl. & Eval  \\
        \midrule
        GPT-4o ($2024$-$08$-$06$)  & $78.5\pm4.1$ & \bm{$69.9\pm6.2$} & $78.1\pm4.9$ & $54.9\pm5.7$ & $92.7$ \\
        o3-mini (high) & \bm{$83.3\pm3.7$} & $66.0\pm6.3$ & \bm{$87.6\pm3.8$} & $59.1\pm5.7$ & - \\
 
        \midrule
        
        Claude $3$ Opus & $75.1\pm4.3$ & $67.4\pm6.2$ & $66.8\pm5.5$ & $69.2\pm5.4$ & $84.9$ \\
        Claude $3.7$ Sonnet (ET.) & $80.9\pm3.9$ & $67.5\pm6.4$ & $86.9\pm4.0$ & $68.5\pm5.4$ & \bm{$97.8$}  \\
        Claude $3.7$ Sonnet & $74.3\pm4.4$ & $57.1\pm6.6$ & $71.0\pm5.3$ & \bm{$69.5\pm5.3$} & $94.9$  \\
        Claude $3.5$ Haiku & $70.9\pm4.5$ & $60.4\pm6.5$ & $66.4\pm5.5$ & $61.2\pm5.6$ & $88.1$  \\
  
        \midrule
  
        Llama $3.3$ $70$B & $58.1\pm4.9$ & $53.7\pm6.8$ & $65.7\pm5.5$ & $49.3\pm5.8$ & $88.4$  \\
        Llama $3.1$ $8$B & $43.7\pm4.9$ & $39.6\pm6.6$ & $36.4\pm5.7$ & $32.8\pm5.4$ & $72.6$  \\
  
        \midrule
  
        DeepSeek-R1 $671$B & $73.3\pm4.4$ & $64.6\pm6.4$ & $77.4\pm4.9$ & $59.4\pm5.6$ & $97.7$ \\
        R1-Distill-Qwen-$14$B & $46.9\pm5.0$ & $38.7\pm6.5$ & $38.9\pm5.7$ & $39.1\pm5.6$ & - \\

        \midrule

        Qwen$2.5$-Coder-$32$B & $70.2\pm4.5$ & $64.7\pm6.4$ & $63.2\pm5.6$ & $56.3\pm5.7$ & $92.1$ \\
        Qwen-QwQ-$32$B & $52.6\pm5.0$ & $47.2\pm6.7$ & $51.6\pm5.8$ & $34.3\pm5.4$ & $97.6$ \\
        
        \bottomrule
    \end{tabular}
    }
    \caption{Overall \BenchP{} and \BenchJ{} results, with HumanEval results. For all open-source models, their ``instruct'' versions are used. ``ET'' refers to the extended thinking version of the model. Unless otherwise specified, Claude $3.7$ Sonnet with ET is used in the following experiments and is simply referred to as Claude $3.7$ Sonnet.}
    \label{tab:llm_comparison}
\end{table}

The Java repositories are all intermediate-to-advanced academic programming projects.  These include: processing a text corpus to build bag-of-words vectors, sparse vector and matrix implementations, generation of random variates, implementation of an AVL tree, a B-tree implementation, and an M-Tree implementation.   
The Python repositories were selected to include notebook-style scripts and more modular, object-oriented codes.  These codes cover classic algorithms such as linear programming, as well as tasks in computer vision and reinforcement learning. See Appendix for examples of Java and Python codes from the benchmark given in Figure \ref{fig:java_processing} and Figure \ref{fig:python_processing}, respectively.

\section{Preparation and Breakdown of Results}

All AI-for-code models are allowed only one chance to produce a result in the \Bench{} benchmark to mimic the real-life application scenario. A user asking an AI to produce a code is likely going to ask one time for a code and either accept (and possibly modify) the result or reject it; it is unlikely that any user is going to look through many different results.  When controllable and applicable, the randomness of the AI-for-code model is ``turned off'' so that the most likely/preferred answer is produced.   
Rather than simply producing one single pass rate, the benchmark produces four pass rates (Java infill and completion; Python infill and completion). The proposed \Bench{} benchmark enables a detailed breakdown of results, which is organized as follows:

\textbf{Pass rate by closest comment distance}.  We compute the distance (number of lines) from the start of the infill/completion task to the closest comment that precedes it.  We measure the overall median distance, and any task whose closest comment distance is less than the median is said to have a ``short'' distance; and any greater than the median is said to have a ``long'' distance. All results in the \Bench{} benchmark are presented with $95\%$ confidence intervals. See Figure \ref{fig:comment_dist_v2}.

\textbf{Pass rate by distance to the furthest referenced program object}.  We compute all named class/function/variable/etc. reference distances in terms of the number of lines of code, and group the references into terciles based on the distance.  A programming task is said to have a ``short'' reference distance if its longest reference distance is in the smallest tercile, a ``medium'' if its longest distance is in the middle tercile, and a ``long'' distance if its longest is in the largest tercile. See Figure \ref{fig:group_by_ref_dist}.

\textbf{Pass rate by program constructs}. We break down the results based on program constructs such as referring to a local or global variable, function, class etc. See Figure \ref{fig:python_construct} and Figure \ref{fig:java_construct}.

\section{Benchmark Results}
\label{sec:bench_result}

\textbf{Experimental Setup.} We ran \Bench{} on a representative set of state-of-the-art large language models (LLMs) for code.  We evaluate a range of both popular closed-source and state-of-the-art open-source language models on our proposed benchmark.

We assess two prominent closed-source model families: the GPT models, including GPT-4o ($2024$-$08$-$06$) \citep{gpt-4o} and O3-mini \citep{o3-mini}, and Claude models from Anthropic, including Claude-$3$-Opus, Claude-$3.5$-Haiku, and Claude-$3.7$-Sonnet \citep{claude-all-models}. We also test Claude $3.7$ Sonnet with extended thinking using a $16$k token budget \citep{claude-3-7-sonnet-ext}. From open-source models, we examine the Llama family, including Llama-$3.3$-$70$B-Instruct \citep{llama3-3} and Llama-$3.1$-$8$B-Instruct \citep{llama3-1}, DeepSeek models \citep{guo2025deepseek} including DeepSeek-R1 and DeepSeek-R1-Distill-Qwen-$14$B, and Qwen models including Qwen$2.5$-Coder-$32$B-Instruct \citep{hui2024qwen2} and Qwen-QwQ-$32$B \citep{qwq32b}. All models operate with their respective intrinsic context lengths. Claude family models and O3-mini use $200$k tokens, while GPT-4o, Llama models, and DeepSeek-R1 employ $128$k tokens. DeepSeek-R1-Distill-Qwen-$14$B and all Qwen-based models utilize $131$k tokens. Table \ref{tab:llm_comparison} gives the overall \Bench{} results for each of the twelve models. For comparison, this table also gives the published HumanEval \citep{chen2021evaluating} performance of each model.  

% \begin{figure}[!ht]
\begin{figure}[h!]
	\centering
	\includegraphics[width= \textwidth]{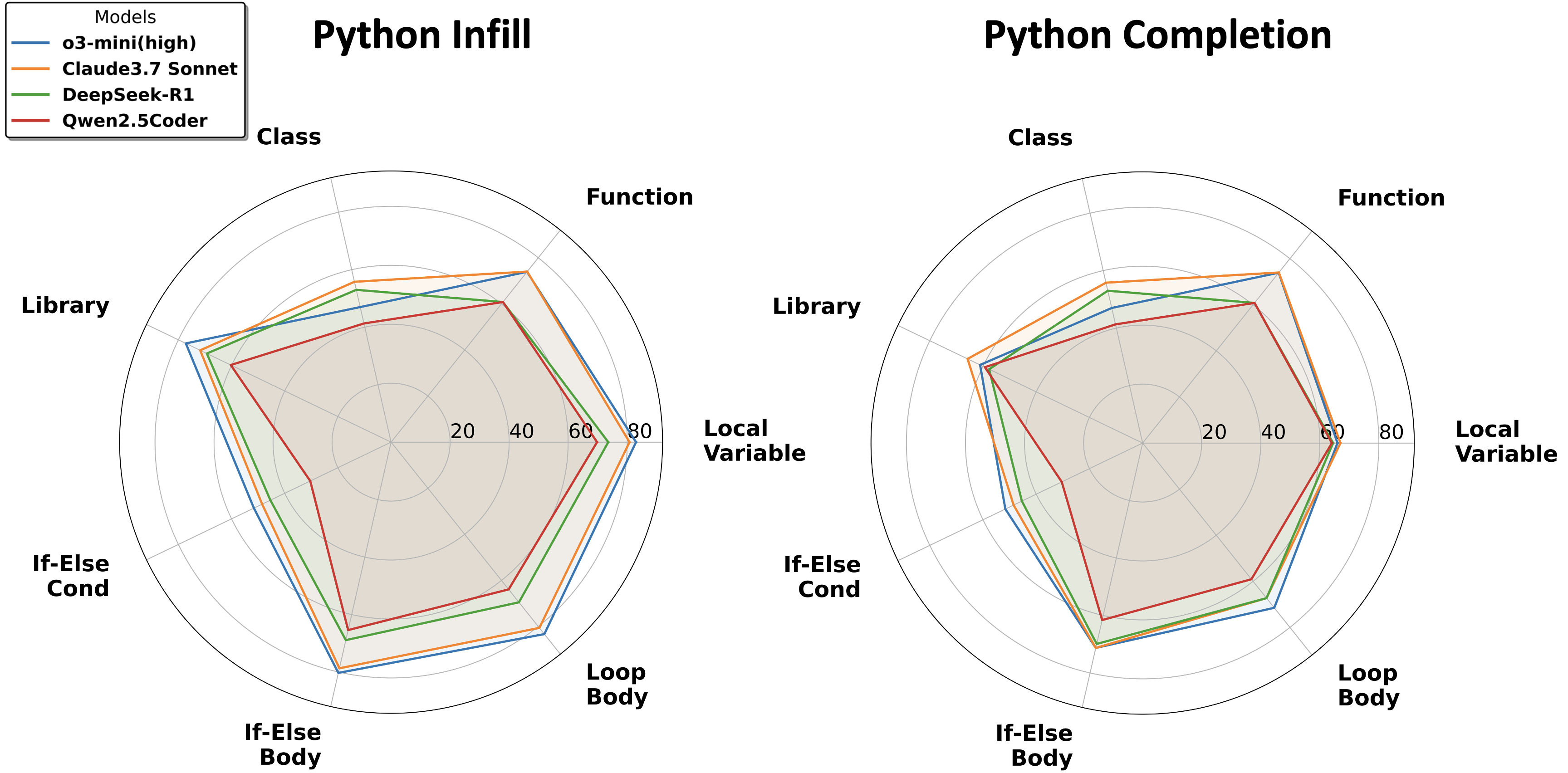}
	\caption{Pass rate group by Python Construct.}
	\label{fig:python_construct}
\end{figure}

% \begin{figure}[!ht]
\begin{figure}[h!]
	\centering
	\includegraphics[width= \textwidth]{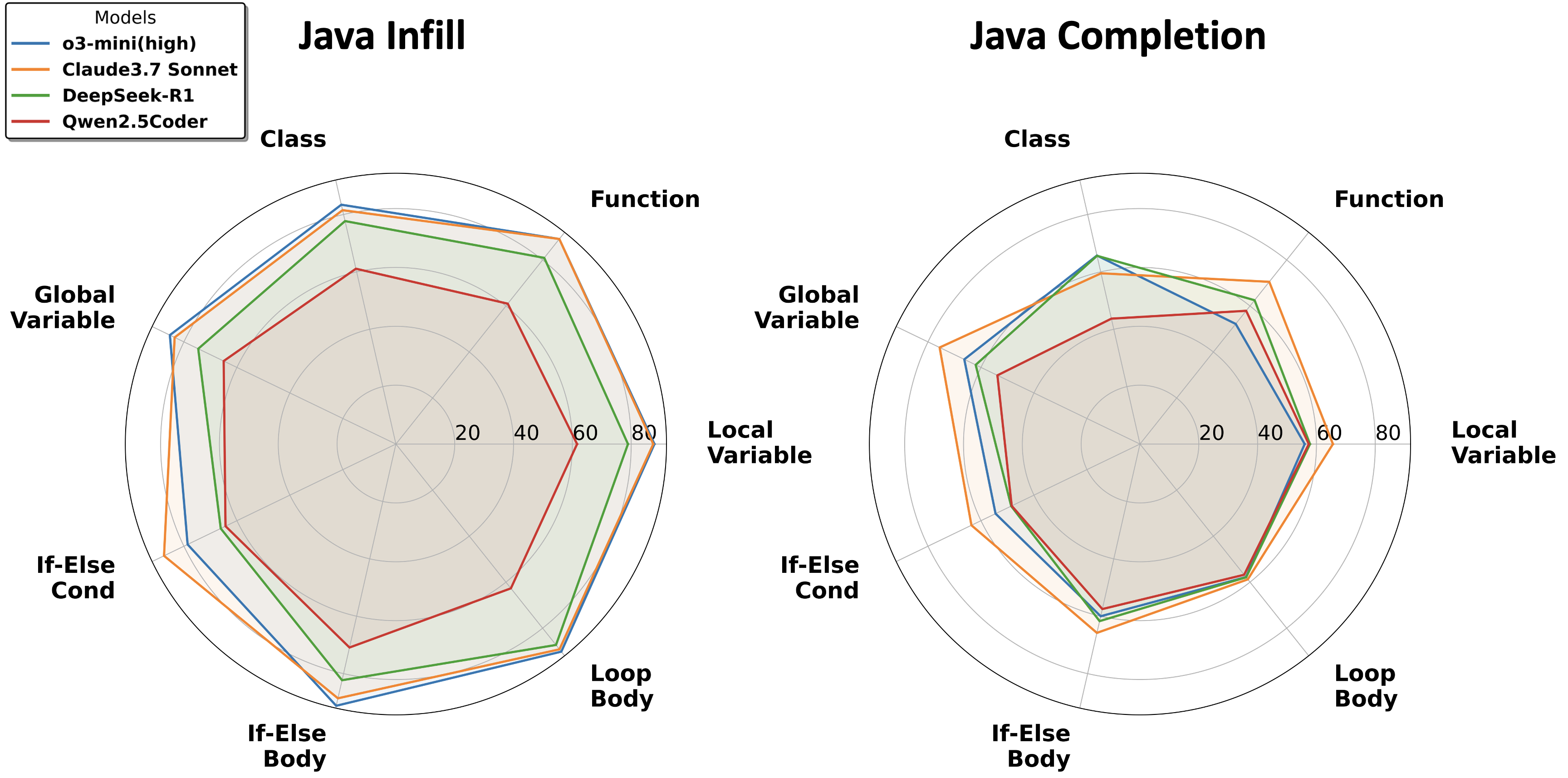}
	\caption{Pass rate grouped by Java Construct.}
	\label{fig:java_construct}
\end{figure}

\textbf{Model Performance Disparities.} As mentioned in the introduction to the paper, one of the most striking findings is that while the various AI-for-code models did generally perform according to expectation in terms of which models scored highest (o3-mini and Claude $3.7$ Sonnet) and lowest (Llama $3.1$ $8$B), the gap between these models was very substantial according to \Bench{}. For example, o3-mini succeeded on $87.6\%$ of Java infill tasks, whereas Llama $3.1$ $8$B succeeded on only $36.4\%$of Java infill tasks—a gap of over $50$ percentage points. At the same time, there was a relatively small gap between these models according to HumanEval. For instance, Claude $3.7$ Sonnet (ET) scored $97.8\%$ on HumanEval while even the lowest performer, Llama $3.1$ $8$B, still achieved $72.6\%$—a much narrower gap of approximately $25$ percentage points. There are several explanations for this discrepancy, but the most obvious is that at this point, many AI-for-code models have likely been trained on HumanEval, and so solving those tasks is more of a recall problem than a genuine programming problem.

\textbf{Context-Driven Advantage: Infill Outperforms Completion.} Our results confirmed the expected difficulty gradient between task types. The largest models (O3-mini and Claude $3.7$ Sonnet with extended thinking) consistently performed better on infill tasks ($83.3-87.6\%$ success rates) than on completion tasks ($59.1-69.5\%$). This pattern makes intuitive sense: code following a missing block provides valuable contextual clues about the required implementation. Completion tasks present inherently greater challenges, requiring models to anticipate programmer intent without subsequent code as guidance. Despite this difficulty, leading models achieve respectable performance with clear instructions—particularly GPT-4o in Python ($69.9\%$) and Claude $3.7$ Sonnet in Java ($69.5\%$). As Figures \ref{fig:python_construct} and \ref{fig:java_construct} reveal, while Java infill performance approaches perfection for the best models across most constructs, completion performance degrades in comparison. This performance difference highlights critical areas for improvement for future programming assistants.

\textbf{Programming Construct Challenges.}
Examining Figure \ref{fig:python_construct} and \ref{fig:java_construct}, several unexpected patterns emerge in model performance across different programming constructs. One surprising finding is the substantial gap between if-else conditions and bodies. Models consistently struggle with generating if-else conditions (Python: $30-52\%$ pass rates) compared to if-else bodies (Python: $65-80\%$). This pattern persists in Java, though less extreme. This suggests that logical conditions require more precise reasoning with fewer explicit clues than implementation bodies, which benefit from clearer structural hints. A second unexpected result is Java models' superior performance on global variable tasks ($75-85\%$) compared to local variables ($56-87\%$), contradicting conventional programming intuition that local scope is simpler. This may be because global variables provide clearer dependencies through explicit references, while local variables introduce complexity through nested scopes and implicit context.

Python class-related tasks also show surprisingly lower accuracy ($41-55\%$) than function tasks ($60-74\%$), despite both being common abstractions. This Python-specific difficulty likely stems from Python's flexible structure providing less explicit context for class definitions, while Java's verbosity offers clearer syntactic cues. Finally, loop body tasks consistently outperform conditional logic and class-related tasks across both languages ($60-90\%$ pass rates), which may arise from loops' inherently repetitive structure providing stronger pattern recognition cues compared to open-ended logical reasoning tasks.

% \begin{figure}[!ht]
\begin{figure}[h!]
	\centering
	\includegraphics[width=0.9\textwidth]{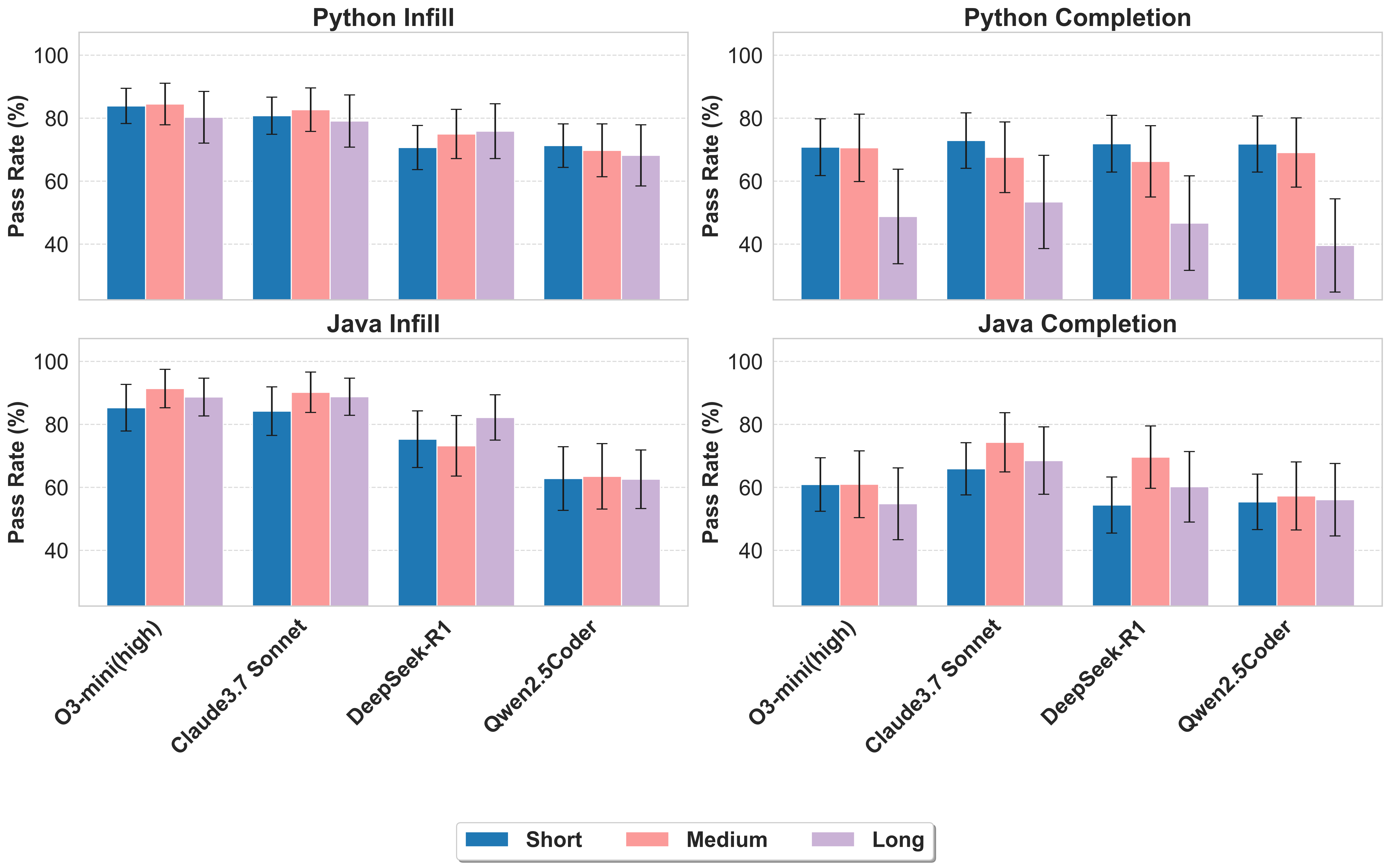}
	\caption{Pass rate grouped by reference object distance.}
	\label{fig:group_by_ref_dist}
\end{figure}

\textbf{Reference Distance Effects.}
A revealing trend emerges when analyzing pass rates by reference-object distance (Figure~\ref{fig:group_by_ref_dist}). For Python completion tasks, performance declines as distance to referenced objects increases, aligning with lower pass rates for complex, long-distance constructs (e.g., class definitions) compared to short-distance tasks (e.g., loop bodies). Interestingly, Java completion shows the opposite pattern: longer-distance references (e.g., global variables, classes) often yield higher pass rates than closer-scoped items (e.g., local variables). This suggests Java's explicit syntactic structure may help models navigate distant dependencies more effectively than Python's more flexible approach.

% \begin{figure}[!ht]
\begin{figure}[h!]
	\centering
	\includegraphics[width=0.9\textwidth]{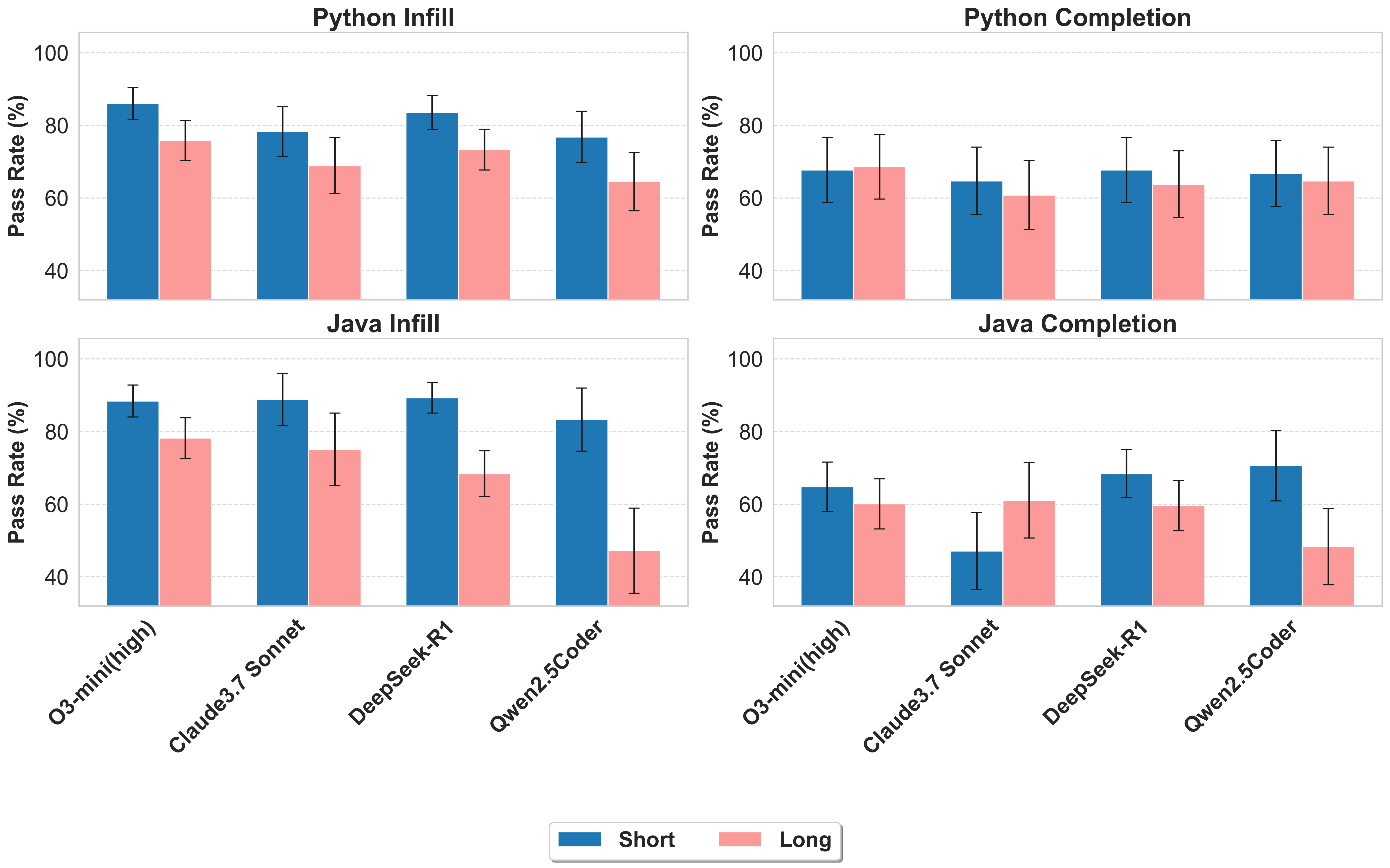}
	\caption{Pass rate grouped by distance to the nearest comment.}
	\label{fig:comment_dist_v2}
\end{figure}

\textbf{Critical Role of Comments.}
Existing benchmarks like MBPP and HumanEval feature comments immediately preceding target code blocks, artificially boosting generation performance and obscuring models' true code-writing capabilities. In real-world development, such dense annotation is impractical and time-consuming. Our benchmark demonstrates that as the distance to the nearest comment increases, code generation quality consistently deteriorates (Figure \ref{fig:comment_dist_v2}), revealing limitations that standard benchmarks might mask.

\section{Conclusion}
\label{sec:conclusion}
In this paper, we have described \Bench{}, which is a benchmark designed to evaluate the ability of an AI-for-code model to perform well in a simulated, Copilot-style environment, where the goal is to supply a programmer with a small bit of missing code in the context of a complex programming project.  \Bench{} considers both infill and completion tasks, and not only returns high-level accuracy numbers, but it also returns fine-grained benchmark results, stratified according to metrics such as the type of code to produce, the distance of the declaration of a program element (such as a global variable) to where it is used, and the extent to which comments are available locally.
\Bench{} will be open-source, and we hope that over time, others will add high-quality, Java and Python infill tasks (not available elsewhere on the web) to \Bench{}, as well as model-specific pre-processing and post-processing codes to the project. 

The most obvious limitation of \Bench{} as it is currently designed is that the numbers produced, while useful, may not accurately reflect user satisfaction with the output of an AI-for-code model.  \Bench{} reports the fraction of codes produced that are correct, in the sense that they pass all test cases.  However, a real-life user may not require correctness, and may often be willing to change a variable name (or two) in a produced code.  Further, \Bench{} exclusively evaluates the ability of an AI-for-code tool to complete an almost-completed, medium-to-large project.  In practice, most projects will not be complete at the time that the code is added, and so there may not be a single correct answer---based on the output of a model (whatever it is), a user may be willing to tailor later code to match the output.  It may be somewhat easier to produce \emph{usable} code than what \Bench{}'s results would suggest.  Still, despite these limitations, we believe that \Bench{} is a useful means of evaluating AI-for-code models. 

% \textbf{Broader Impact and Ethics Statement.} While \Bench{} promotes AI-for-code precision, enhancing development efficiency, developers might over-rely on these tools, diminishing human oversight in coding.
% \input{text/data_availability}

\newpage
\small{
\bibliographystyle{colm2025_conference}
\bibliography{colm2025_conference}
}
\newpage

%%%%%%%%%supplement%%%%%%%%%%%
\newpage
\appendix
\section*{\centering \LARGE \textbf{\Bench{}: Evaluating Large Language Models for Copilot-Style Code Generation Appendix}}
\addcontentsline{toc}{section}{\Bench{}: Evaluating Models for Copilot-Style Code Generation Appendix}

\section{Details of \Bench{}}
\textbf{Dataset Documentation and Intended Uses.}
This dataset, compiled from eight Java and seven Python code repositories, is designed primarily for the evaluation of AI-for-code models. It can also be utilized responsibly for educational purposes, software engineering studies, and the development of new coding tools. 

\textbf{Data Collection.}
Emails were sent to faculty and students within the University's Computer Science, Electrical Engineering, and Statistics departments, inviting them to contribute Java and Python private code repositories for AI-for-code research. Upon receipt, $1,163$ code generation tasks were curated to ensure a diverse and representative sample of real-world code, gathering approximately $11,000$ lines of code.

\textbf{Code Block Extraction.} For every sourced code repository, the following Algorithm was used to extract self-contained code blocks - 
\begin{algorithm}
\caption{Selecting Logically Self-Contained Code Blocks}
\begin{algorithmic}[1]
\STATE blocks = [] \hfill $\triangleright$ Initialize an empty list to store code blocks with starting and end points.
\STATE ProgConstructList = [Loop, If-Else, Function] \hfill $\triangleright$ Define constructs of interest.
\FOR{file in files}
    \STATE ast = parse(file) \hfill $\triangleright$ Parse the file using AST parser.
    \FOR{node in ast}
        \IF{node.type in ProgConstructList}
            \STATE DFS\_traverse\_nodes(node, blocks) \hfill $\triangleright$ Perform DFS to traverse the node body.
        \ENDIF
    \ENDFOR
\ENDFOR
\STATE return blocks \hfill $\triangleright$ Return the list of code blocks with starting and end points.
\end{algorithmic}
\label{alg:annotation}
\end{algorithm}

% \begin{algorithm}
% \caption{DFS\_traverse\_nodes(node, blocks)}
% \begin{algorithmic}[1]
% \STATE start = node.start \hfill $\triangleright$ Set the starting point at the beginning of the node.
% \STATE end = node.end \hfill $\triangleright$ Initialize the end point at the end of the node.
% \STATE blocks.append((start, end)) \hfill $\triangleright$ Add the code block (start, end) to the list.
% \FOR{child in node.body}
%     \IF{child.type in ProgConstructList}
%         \STATE DFS\_traverse\_nodes(child, blocks) \hfill $\triangleright$ Recursively traverse the child node.
%     \ENDIF
% \ENDFOR
% \end{algorithmic}
% \end{algorithm}

Below, we further discuss the additional details of extracting/selecting the code blocks -
\begin{itemize}
    \item \textbf{Nested Constructs:} To ensure logical self-containment, the selection of code blocks starts at the beginning of a construct (e.g., loop body) and ends at the end of the same construct, even if it includes nested constructs such as if statements within the loop. This approach maintains the logical integrity and completeness of the selected code block.
    
    \item \textbf{Completion vs. Infill Tasks:} The selection criteria differ between completion and infill tasks due to their distinct contexts. Infill tasks have both preceding and succeeding code, providing context that helps the model generate code that fits seamlessly into the existing sequence. In contrast, completion tasks only have preceding code and lack postceding context, making it essential to define clear stopping points.  Without precise stopping points, ambiguity arises, leading to potential duplication or conflict when the generated code is combined with subsequent blocks. By focusing on the nearest incomplete construct, such as the end of a statement or function, the model can avoid extending beyond a well-defined boundary.
\end{itemize}

\textbf{Human Annotation.}
To ensure high-quality code generation tasks, a team of graduate students from the University, each possessing $5$ to $10$ years of programming experience and ranging from medium to advanced skill levels, meticulously analyzed the candidate code blocks returned by Algorithm \ref{alg:annotation} for each source code. The generated coding tasks encompass both infill and completion types, distributed across eight distinct program categories. A detailed statistical breakdown of the tasks within each category is provided in Table \ref{tab:task-freq}. Annotators followed a shared rule-based protocol for block selection and task creation. In cases of disagreement, we resolved conflicts via majority vote and excluded tasks where consensus could not be reached. All selected blocks were reviewed to ensure structural integrity, syntactic correctness, and task relevance.

% \textbf{Dataset and Metadata Access.} The dataset and its associated metadata, is hosted on an anonymous google drive  available through the below link
% \url{https://anonymous.4open.science/r/SimCoPilot-0058}.

\newpage
\textbf{Hosting, Licensing, and Maintenance Plan.}
\begin{itemize}
    \item \textbf{Hosting Platform:} The dataset and its associated metadata, documented using the Croissant metadata framework, can be viewed and downloaded at \url{https://huggingface.co/datasets/mj33/SimCoPilot}. Code and associated files are also available through the following link: \href{https://github.com/mj33rice/Sim-CoPilot}{https://github.com/mj33rice/Sim-CoPilot}.
    \item \textbf{Maintenance Plan:} We commit to maintaining the dataset with regular updates and revisions to correct any issues and integrate new contributions. Updates will be documented in the repository's release notes section.
    \item \textbf{Licensing:} The data is shared under the [CC BY-NC-ND 4.0] and code is licensed under MIT License.
\end{itemize}

\textbf{Legal and Ethical Responsibility.}
The authors of this dataset bear full responsibility for any violation of rights.

\section{Details of Post-Processing Steps}
Post-processing is essential as even the best AI-for-code models can generate superfluous English comments, repeat lines, or entire blocks of code, often contrary to explicit instructions in the prompts. To address these issues, models are equipped with a specialized post-processor. The \Bench{} benchmark incorporates such a post-processor, which performs several crucial steps to refine the output:

\begin{enumerate}
    \item \textbf{Remove Non-Code Syntax:} using a series of regular expressions to eliminate any text that is clearly not a statement in the target programming language.
    \item \textbf{Remove Duplicate Code:} Removes duplicated code that appears above the prompt in completion tasks or following the prompt in infill tasks.
    \item \textbf{Auto Indentation or Brackets Correction:} Adjusts indentation for Python and corrects bracket placement for Java, which are critical for maintaining the structural integrity of the code.
\end{enumerate}

\begin{figure}[t!]
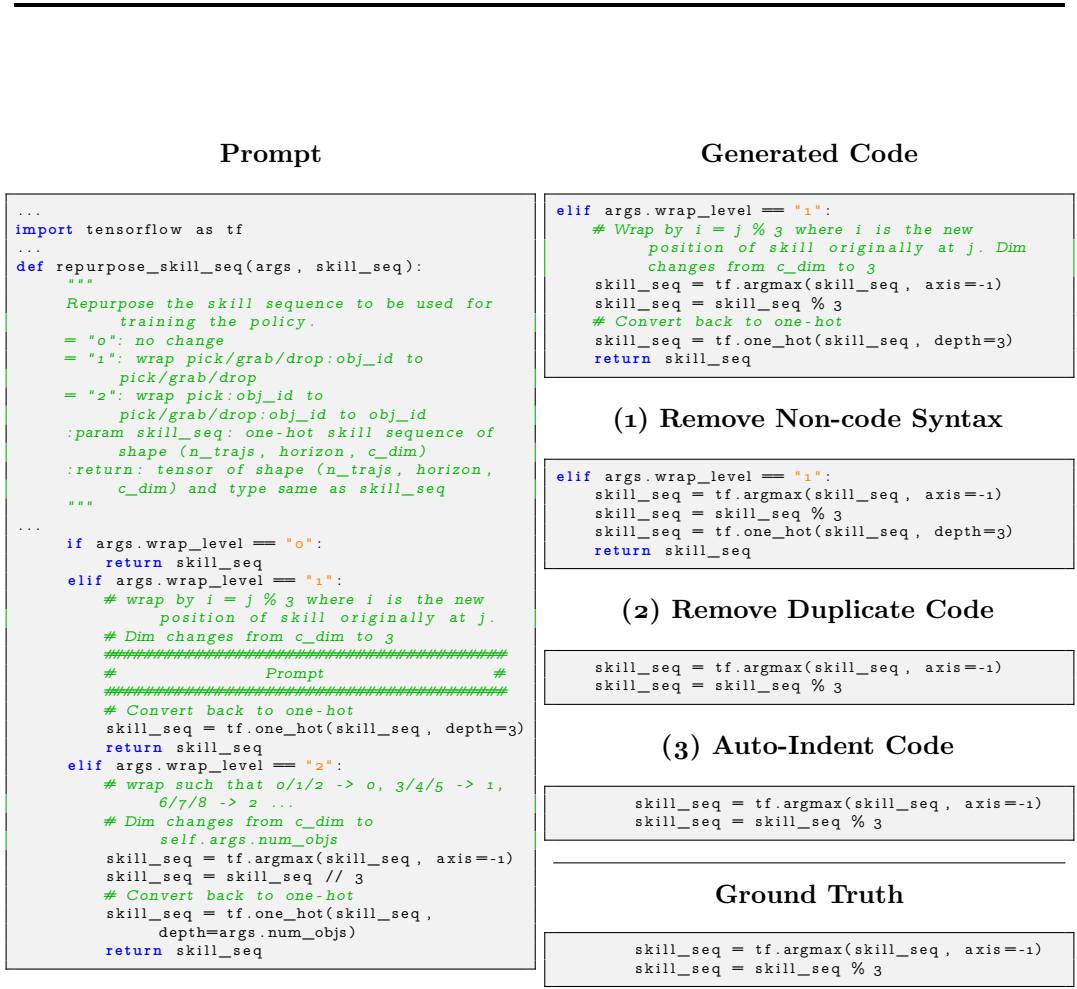

    \centering
    \begin{multicols}{2}
        % Left column
        \begin{minipage}[t]{\linewidth}
            \vspace{0pt}
            % \rule{\columnwidth}{0.4pt}
            \begin{center}
                \textbf{Prompt}
            \end{center}
\begin{lstlisting}[style=customPython]
...
import tensorflow as tf
...
def repurpose_skill_seq(args, skill_seq):
     """
     Repurpose the skill sequence to be used for training the policy.
     = "0": no change
     = "1": wrap pick/grab/drop:obj_id to pick/grab/drop
     = "2": wrap pick:obj_id to pick/grab/drop:obj_id to obj_id
     :param skill_seq: one-hot skill sequence of shape (n_trajs, horizon, c_dim)
     :return: tensor of shape (n_trajs, horizon, c_dim) and type same as skill_seq
     """
...
     if args.wrap_level == "0":
         return skill_seq
     elif args.wrap_level == "1":
         # wrap by i = j % 3 where i is the new position of skill originally at j. 
         # Dim changes from c_dim to 3
         ########################################
         #               Prompt                 # 
         ########################################
         # Convert back to one-hot
         skill_seq = tf.one_hot(skill_seq, depth=3)
         return skill_seq
     elif args.wrap_level == "2":
         # wrap such that 0/1/2 -> 0, 3/4/5 -> 1, 6/7/8 -> 2 ... 
         # Dim changes from c_dim to self.args.num_objs
         skill_seq = tf.argmax(skill_seq, axis=-1)
         skill_seq = skill_seq // 3
         # Convert back to one-hot
         skill_seq = tf.one_hot(skill_seq, depth=args.num_objs)
         return skill_seq
\end{lstlisting}
        \end{minipage}
        
        % Right column
        \begin{minipage}[t]{\linewidth}
            \vspace{0pt}
            % \rule{\columnwidth}{0.4pt}
            \begin{center}
                \textbf{Generated Code}
            \end{center}
\begin{lstlisting}[style=customPython]
elif args.wrap_level == "1":
    # Wrap by i = j % 3 where i is the new position of skill originally at j. Dim changes from c_dim to 3
    skill_seq = tf.argmax(skill_seq, axis=-1)
    skill_seq = skill_seq % 3
    # Convert back to one-hot
    skill_seq = tf.one_hot(skill_seq, depth=3)
    return skill_seq
\end{lstlisting}
            
            \vspace{0pt}
            % \rule{\columnwidth}{0.4pt}
            \begin{center}
                \textbf{(1) Remove Non-code Syntax}
            \end{center}
\begin{lstlisting}[style=customPython]
elif args.wrap_level == "1":
    skill_seq = tf.argmax(skill_seq, axis=-1)
    skill_seq = skill_seq % 3
    skill_seq = tf.one_hot(skill_seq, depth=3)
    return skill_seq
\end{lstlisting}
            
            \vspace{0pt}
            % \rule{\columnwidth}{0.4pt}
            \begin{center}
                \textbf{(2) Remove Duplicate Code}
            \end{center}
\begin{lstlisting}[style=customPython]
    skill_seq = tf.argmax(skill_seq, axis=-1)
    skill_seq = skill_seq % 3
\end{lstlisting}
            
            \vspace{0pt}
            % \rule{\columnwidth}{0.4pt}
            \begin{center}
                \textbf{(3) Auto-Indent Code}
            \end{center}
\begin{lstlisting}[style=customPython]
        skill_seq = tf.argmax(skill_seq, axis=-1)
        skill_seq = skill_seq % 3
\end{lstlisting}

            \vspace{-5pt}
            \rule{\columnwidth}{0.4pt}
            \begin{center}
                \textbf{Ground Truth}
            \end{center}
\begin{lstlisting}[style=customPython]
        skill_seq = tf.argmax(skill_seq, axis=-1)
        skill_seq = skill_seq % 3
\end{lstlisting}
        \end{minipage}
    \end{multicols}
    \caption{Example infill task from \BenchP{} and step-by-step post-processing demonstration}
    \label{fig:python_processing}
\end{figure}

\begin{figure}[t!]
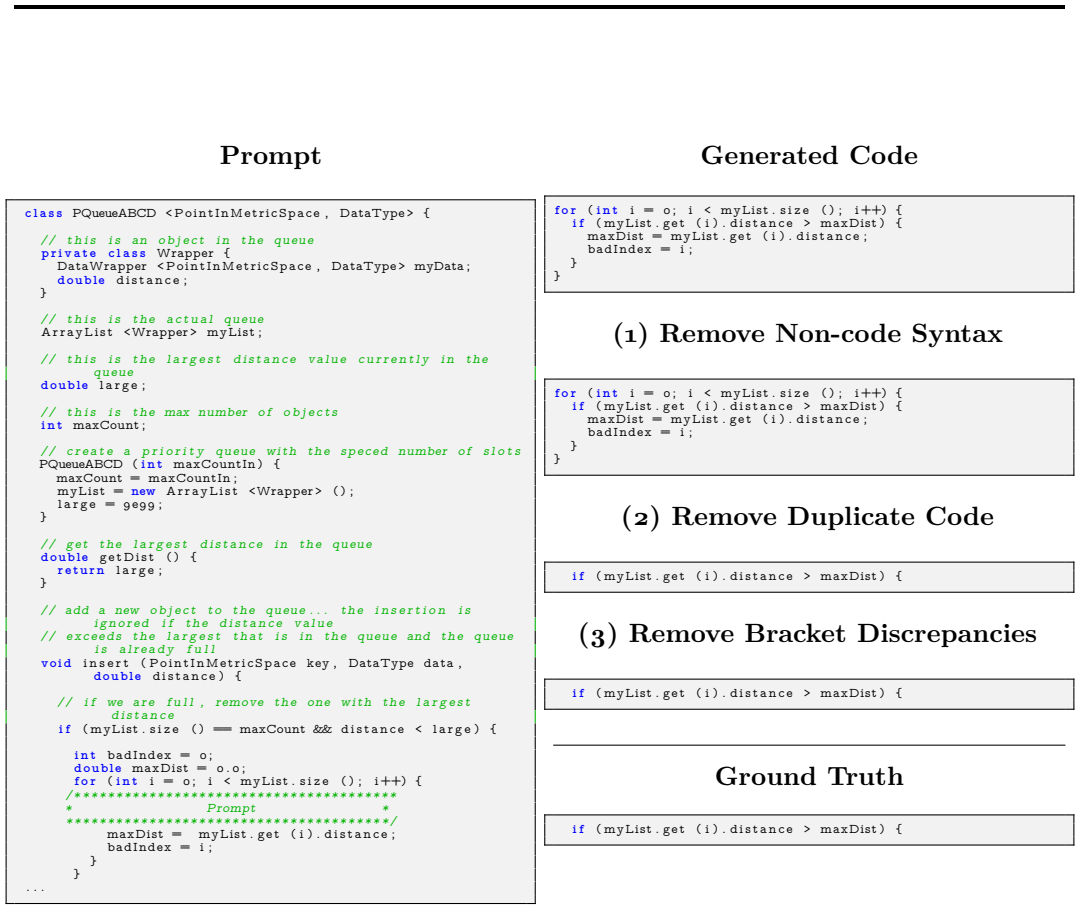

    \centering
    \begin{multicols}{2}
        % Left column
        \begin{minipage}[t]{\linewidth}
        % \begin{minipage}[t]{1.1\linewidth}
            \vspace{0pt}
            % \rule{\columnwidth}{0.4pt}
            \begin{center}
                \textbf{Prompt}
            \end{center}
\begin{lstlisting}[style=customJava]
 class PQueueABCD <PointInMetricSpace, DataType> {

   // this is an object in the queue
   private class Wrapper {
     DataWrapper <PointInMetricSpace, DataType> myData;
     double distance;
   }

   // this is the actual queue
   ArrayList <Wrapper> myList;

   // this is the largest distance value currently in the queue
   double large;

   // this is the max number of objects
   int maxCount;

   // create a priority queue with the speced number of slots
   PQueueABCD (int maxCountIn) {
     maxCount = maxCountIn;
     myList = new ArrayList <Wrapper> ();
     large = 9e99;
   }

   // get the largest distance in the queue
   double getDist () {
     return large;
   }

   // add a new object to the queue... the insertion is ignored if the distance value
   // exceeds the largest that is in the queue and the queue is already full
   void insert (PointInMetricSpace key, DataType data, double distance) {

     // if we are full, remove the one with the largest distance
     if (myList.size () == maxCount && distance < large) {

       int badIndex = 0;
       double maxDist = 0.0;
       for (int i = 0; i < myList.size (); i++) {
			/***************************************
  		*                Prompt               * 
			***************************************/
           maxDist =  myList.get (i).distance;
           badIndex = i;
         }
       }
 ...
\end{lstlisting}
        \end{minipage}
        
        % Right column
        % \begin{minipage}[t]{0.8\linewidth}
        \begin{minipage}[t]{\linewidth}
            \vspace{0pt}
            % \rule{\columnwidth}{0.4pt}
            \begin{center}
                \textbf{Generated Code}
            \end{center}
\begin{lstlisting}[style=customJava]
for (int i = 0; i < myList.size (); i++) {
  if (myList.get (i).distance > maxDist) {
    maxDist = myList.get (i).distance;
    badIndex = i;
  }
}
\end{lstlisting}
            
            \vspace{0pt}
            % \rule{\columnwidth}{0.4pt}
            \begin{center}
                \textbf{(1) Remove Non-code Syntax}
            \end{center}
\begin{lstlisting}[style=customJava]
for (int i = 0; i < myList.size (); i++) {
  if (myList.get (i).distance > maxDist) {
    maxDist = myList.get (i).distance;
    badIndex = i;
  }
}
\end{lstlisting}
            
            \vspace{0pt}
            % \rule{\columnwidth}{0.4pt}
            \begin{center}
                \textbf{(2) Remove Duplicate Code}
            \end{center}
\begin{lstlisting}[style=customJava]
 	if (myList.get (i).distance > maxDist) {
\end{lstlisting}

            \vspace{0pt}
            % \rule{\columnwidth}{0.4pt}
            \begin{center}
                \textbf{(3) Remove Bracket Discrepancies}
            \end{center}
\begin{lstlisting}[style=customJava]
 	if (myList.get (i).distance > maxDist) {
\end{lstlisting}

            \vspace{0pt}
            \rule{\columnwidth}{0.4pt}
            \begin{center}
                \textbf{Ground Truth}
            \end{center}
\begin{lstlisting}[style=customJava]
 	if (myList.get (i).distance > maxDist) {
\end{lstlisting}
        \end{minipage}
    \end{multicols}
    \caption{Example completion task from \BenchJ{} and step-by-step post-processing demonstration}
    \label{fig:java_processing}
\end{figure}

It is important to note that while post-processing adjusts the overall indentation and syntax to the entire generated code block, it preserves the relative indentations within code blocks as the way the model generates. This approach avoids altering the original programming intent, such as the termination points of loops or conditional blocks. For detailed examples of how these post-processing steps are applied in practice, refer to the step-by-step code illustrations for Python infill tasks and Java completion tasks in Figure \ref{fig:python_processing} and Figure \ref{fig:java_processing}, respectively.

% \begin{figure*}[ht!]
%     \centering
%     \begin{subfigure}{\textwidth}
%         \centering
%         \includegraphics[width=0.7\textwidth]{supplement/figs/Java_pass_rate_comparison_V2.png}
%         \caption{Pass rates with versus without Post-Processing for 569 Programming Tasks in \BenchJ{}.}
%         \label{fig:sub1}
%     \end{subfigure}
%     % Adding some vertical space between the subfigures
%     \vspace{1cm}
%     \begin{subfigure}{\textwidth}
%         \centering
%         \includegraphics[width=0.7\textwidth]{supplement/figs/Python_pass_rate_comparison_V2.png}
%         \caption{Pass rates with versus without Post-Processing for 594 Programming Tasks in \BenchP{}.}
%         \label{fig:sub2}
%     \end{subfigure}
%     \caption{Comparison of pass rates with versus without post-processing.}
%     \label{fig:post_process}
% \end{figure*}

It is important to note that in our comparison labeled ``without'' post-processing, the post-processor was still utilized to eliminate non-code syntax, such as explanations or boilerplate text like ``Here is the generated code:'' which the best AI models frequently produce despite instructions to the contrary from the prompt. This step reflects realistic software development practices, where developers do not indiscriminately copy all text generated by AI models into their projects.

% In Figure \ref{fig:post_process}, 
We compare the pass ratios with and without the application of comprehensive post-processing steps on \BenchJ{} and \BenchP{}. The data clearly shows that post-processing markedly boosts the pass rates—doubling it for larger models and increasing it tenfold for smaller models. Notably, the comparative analysis between Python and Java tasks reveals that post-processing yields a significantly greater improvement in pass rates on Java, at least doubling them, even with the most advanced AI models. This finding suggests that, in comparison to Python, Java is more sensitive to syntax and code structure, highlighting the importance of tailored post-processing techniques to enhance model performance in different programming environments.

% Define the custom style for Python code
\lstdefinestyle{customPython}{
    language=Python,
    basicstyle=\tiny\ttfamily,
    keywordstyle=\color{blue}\bfseries,
    commentstyle=\color{green!70!black}\itshape,
    stringstyle=\color{orange},
    backgroundcolor=\color{lightgray!20},
    showspaces=false,
    showstringspaces=false,
    showtabs=false,
    frame=single,
    tabsize=4,
    breaklines=true,
    breakatwhitespace=true,
    morekeywords={your,additional,keywords,here},
}

\newpage
\lstdefinestyle{customJava}{
  language=Java,
  basicstyle=\cmdfontsize\ttfamily,
  % basicstyle=\tiny\ttfamily,
  keywordstyle=\color{blue}\bfseries,
  commentstyle=\color{green!70!black}\itshape,
  stringstyle=\color{orange},
  tabsize=2,
  showspaces=false,
  showstringspaces=false,
  breaklines=true,
  breakatwhitespace=true,
  frame=single,
  backgroundcolor=\color{gray!10},  % Shade of gray that is 10% black and 90% white
  morekeywords={your,additional,keywords,here}
}

\newpage
\section{Error Analysis}

\begin{figure*}[t!]
     \centering
\includegraphics[width=1\textwidth]{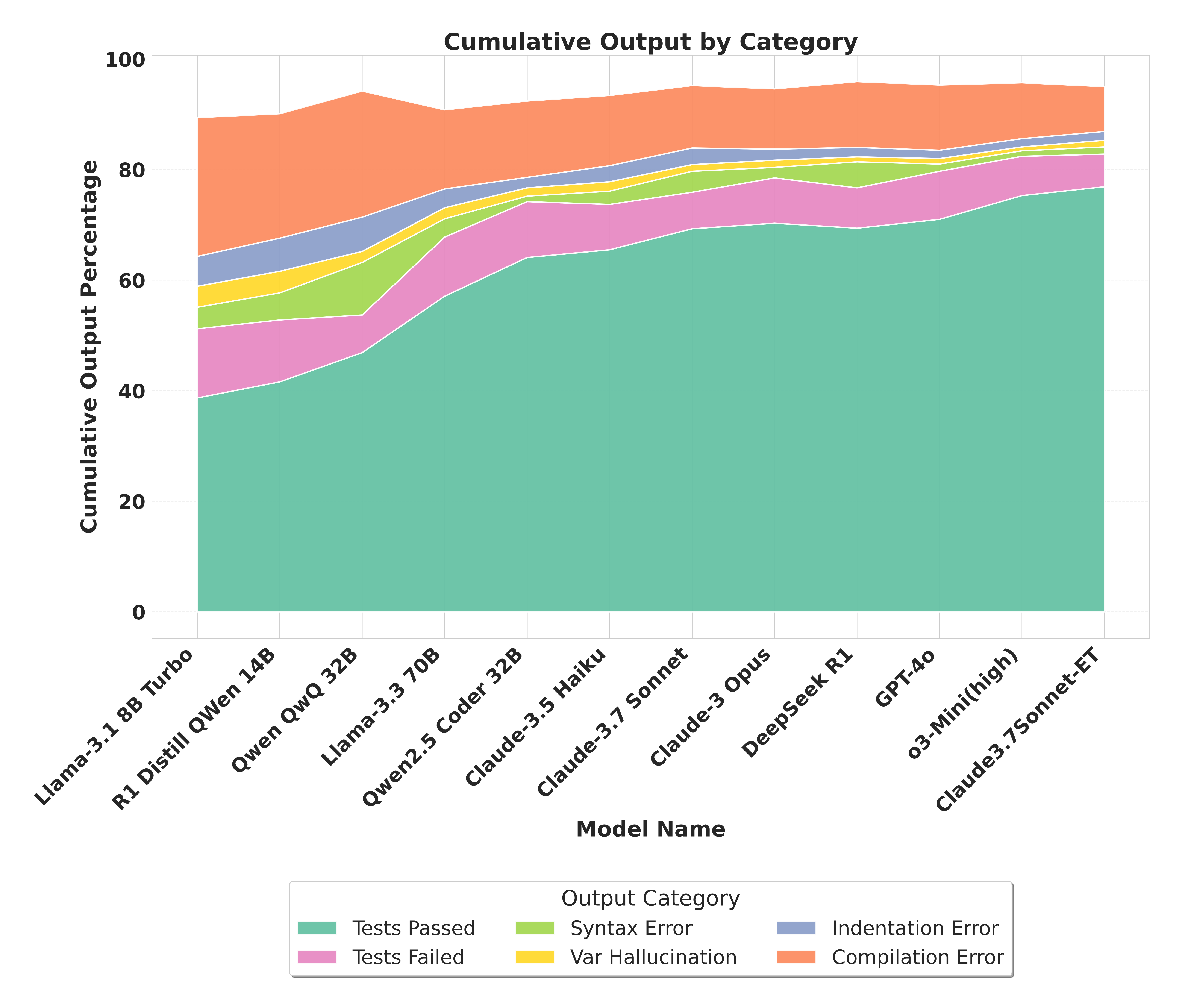}
     \caption{Cumulative Output by Category breakdown for each model among 1,163 programming tasks in \Bench{}. }
     \label{fig:error_analysis}
\end{figure*}

In Figure~\ref{fig:error_analysis}, we analyze the four primary error categories---\textit{Compilation Error}, \textit{Indentation Error}, \textit{Syntax Error}, and \textit{Var Hallucination}---across $1{,}163$ code-generation tasks. We present these errors for each model as a percentage of total outcomes.\\

\textbf{Compilation Error} \textsc{Llama-$3.1$ $8$B Turbo} exhibits the highest compilation error rate at roughly $25\%$. By contrast, \textsc{Claude-$3.7$ Sonnet-ET} stands out with the lowest rate, around $8\%$. The remaining models cluster between these extremes, in the $10-23\%$ range. While larger or more refined models still tend to fare better, no single model is completely immune to compilation failures, suggesting that some misunderstandings of code structure and language rules persist.\\

\textbf{Indentation Error}
Because these models primarily generate Python code in our benchmark, indentation errors remain a distinct category. Overall, indentation errors stay comparatively low in the new results, topping out at about $6\%$ for models like \textsc{Qwen-QwQ-$32$B}, while \textsc{Claude-$3.7$ Sonnet-ET} and \textsc{GPT} family rarely produce indentation issues (near $1.5\%$). This consistency suggests that the majority of systems have developed fairly robust handling of Python’s indentation rules.\\

\textbf{Syntax Error}
Syntax errors follow a clear size‐based trend, with larger or more advanced models displaying fewer syntactic mistakes. The worst cases (e.g.\ \textsc{Qwen-QwQ-$32$B}) approach $9.5\%$ syntax error rates, whereas \textsc{GPT-4o} and \textsc{o3-Mini(high)} lead at around $1\%$. These findings reinforce the notion that stronger language‐model pretraining and instruction tuning correlate with fewer raw syntax violations.\\

\textbf{Var Hallucination}
Variable Hallucination primarily arises from using undefined identifiers or mishandling scope. The best models (e.g.\ \textsc{o3-Mini(high)} and \textsc{Claude-$3.7$ Sonnet-ET}) achieve hallucination rates of about $1\%$, whereas smaller or less consistent models can climb to around $4\%$. This error type remains a strong indicator of how well a model tracks variable definitions and maintains context over longer sequences, a capacity only loosely correlated with model size.\\

\textbf{General Observations.}
 Across the board, syntax and compilation errors remain the two most prevalent error types. These patterns suggest that, while code‐generation models have substantially improved, structural misunderstandings and incomplete coverage of language rules continue to pose significant challenges.

\section{Additional Evaluation Results and Ablations}

Beyond the evaluations of O3-mini, Claude $3.7$ Sonnet with extended thinking, DeepSeek-R1 $671$B, Qwen$2.5$ Coder-$32$B presented in Table \ref{tab:llm_comparison}, this section details results for eight additional models: Claude $3.7$ Sonnet, Claude $3.5$ Haiku, Llama $3.3$ $70$B, Llama $3.1$ $8$B,DeepSeek-R1-Distill-Qwen-$14$B, Qwen-QwQ-$32$B, and GPT-4o. These models were assessed using the same experimental setups described in the main paper, which include features such as distance to the nearest referenced object, proximity to the nearest comment, and various task types. These tasks incorporate different programming constructs, including references to local variables, functions, and loop bodies, allowing for a comprehensive analysis of model performance across diverse coding scenarios.
\newpage
\textbf{Pass rate by distance to the furthest referenced program object}.\\
As observed with the four models described in the main paper, the influence of the distance to referenced methods, functions, and variables on the pass rate of the generated code was not consistently significant. According to Tables \ref{tab:Length_Dep_results_combined}, it is challenging to discern any consistent pattern regarding how program distance affects accuracy. For Python completion tasks, greater distances tended to correlate with increased difficulty, which aligns with expectations. Interestingly, for Java infill tasks, greater distances appeared to actually simplify the task. However, despite these general trends, there are one or two models, as outlined in the tables, that deviate from these patterns.

\textbf{Pass rate by closest comment distance}.\\
We filtered out tasks with no comment above and ended with $207$ tasks for Python completion, $374$ tasks for Python infilling, $278$ tasks for Java completion, and $278$ tasks for Java infilling. It can be observed that comments located near the target code segments were distinctly beneficial. As illustrated in Table \ref{tab:comments_combined}, having comments close to the areas of code to be completed, markedly improved the pass rate of both infill and completion tasks.

% \newpage
\textbf{Pass rate by task type}
We further provide pass rates stratified by the task type for the remainder of models in Tables \ref{tab:by-construct1}, \ref{tab:by-construct2}, \ref{tab:by-construct3}.

\begin{table}[t]
  \centering
  \captionsetup{skip=5pt}
  \centerline{
  \begin{tabular}{cccccc}
    \toprule
    & Distance & O3-mini & C$3.7$ Sonnet* & DeepSeek-R1 & Qwen$2.5$Coder\\
    \midrule
    Infill & Short & {$83.9\pm5.6$} & $80.8\pm5.9$ & $70.7\pm7.0$ & $71.3\pm6.9$ \\
    Python & Med & {$84.5\pm6.6$} & $82.7\pm6.9$ & $75.0\pm7.8$ & $69.8\pm8.4$ \\
           & Long & {$80.3\pm8.2$} & $79.1\pm8.3$ & $75.9\pm8.7$ & $68.2\pm9.7$ \\
    \midrule
    Compl. & Short & $70.9\pm9.1$ & {$72.9\pm8.8$} & $71.9\pm9.0$ & $71.8\pm8.9$ \\
    Python & Med & {$70.6\pm10.9$} & $67.6\pm11.2$ & $66.3\pm11.3$ & $69.1\pm11.0$ \\
           & Long & $49.0\pm14.9$ & {$53.4\pm14.8$} & $46.7\pm15.0$ & $39.6\pm14.8$ \\
    \midrule                                
    Infill & Short & {$85.3\pm7.4$} & $84.2\pm7.7$ & $75.3\pm9.0$ & $62.8\pm10.1$ \\
    Java   & Med & {$91.4\pm6.1$} & $90.2\pm6.4$ & $73.2\pm9.6$ & $63.5\pm10.4$ \\
           & Long & $88.7\pm6.0$ & {$88.8\pm5.9$} & $82.2\pm7.2$ & $62.6\pm9.3$ \\
    \midrule
    Compl. & Short & $60.9\pm8.5$ & {$65.9\pm8.3$} & $54.4\pm8.9$ & $55.4\pm8.8$ \\
    Java   & Med & $61.0\pm10.6$ & {$74.3\pm9.4$} & $69.6\pm9.9$ & $57.3\pm10.8$ \\
           & Long & $54.8\pm11.4$ & {$68.5\pm10.7$} & $60.2\pm11.2$ & $56.1\pm11.5$ \\
    \midrule
    & & GPT-4o & C$3$ Opus & C$3.7$ Sonnet & C$3.5$ Haiku\\
    \midrule
    Infill & Short & {$80.8\pm6.0$} & $77.8\pm6.3$ & $74.2\pm6.7$ & $71.8\pm6.9$ \\
    Python & Med & {$76.7\pm7.7$} & $74.1\pm8.0$ & $73.3\pm8.1$ & $75.0\pm7.8$ \\
           & Long & $75.8\pm8.8$ & $71.4\pm9.3$ & {$75.9\pm8.9$} & $64.9\pm9.6$ \\
    \midrule
    Compl. & Short & {$77.1\pm8.3$} & $72.8\pm8.9$ & $66.6\pm9.6$ & $69.7\pm9.2$ \\
    Python & Med & {$69.1\pm11.1$} & $67.7\pm11.1$ & $58.8\pm11.7$ & $60.3\pm11.6$ \\
           & Long & $53.5\pm15.1$ & {$53.6\pm15.0$} & $32.6\pm13.9$ & $37.4\pm14.3$ \\
    \midrule
    Infill & Short & {$78.6\pm8.6$} & $67.4\pm9.8$ & $64.1\pm10.0$ & $74.1\pm9.1$ \\
    Java   & Med & $76.8\pm9.1$ & $66.9\pm10.2$ & {$78.1\pm9.1$} & $69.5\pm10.1$ \\
           & Long & {$77.5\pm7.9$} & $67.3\pm8.8$ & $72.9\pm8.4$ & $59.8\pm9.4$ \\
    \midrule
    Compl. & Short & $53.6\pm8.9$ & {$72.3\pm7.8$} & $67.4\pm8.3$ & $63.4\pm8.5$ \\
    Java   & Med & $59.8\pm10.6$ & $69.3\pm10.0$ & {$75.7\pm9.3$} & $70.8\pm9.9$ \\
           & Long & $50.7\pm11.4$ & $63.0\pm11.1$ & {$67.1\pm10.8$} & $45.2\pm11.4$ \\
    \midrule
    & & Llama$3.3$ $70$B & Qwen-QwQ-$32$B & R1-Qwen $14$B & Llama$3.1$ $8$B \\
    \midrule
    Infill & Short & $63.5\pm7.3$ & {$52.7\pm7.6$} & $47.8\pm7.5$ & $50.3\pm7.5$ \\
    Python & Med & $50.0\pm9.2$ & {$51.7\pm9.1$} & $45.8\pm9.0$ & $33.7\pm8.6$ \\
           & Long & $60.5\pm10.0$ & {$54.9\pm10.3$} & $44.0\pm10.1$ & $44.0\pm10.2$ \\
    \midrule
    Compl. & Short & $58.4\pm10.0$ & {$57.3\pm10.0$} & $40.7\pm9.8$ & $43.8\pm10.0$ \\
    Python & Med & $57.3\pm11.7$ & {$42.7\pm11.7$} & $48.6\pm12.0$ & $41.1\pm11.8$ \\
           & Long & $37.2\pm14.5$ & {$30.2\pm13.8$} & $16.3\pm11.0$ & $23.2\pm12.5$ \\
    \midrule
    Infill & Short & $60.6\pm10.1$ & {$51.7\pm10.5$} & $42.7\pm10.3$ & $40.4\pm10.1$ \\
    Java   & Med & $65.8\pm10.3$ & {$45.0\pm10.8$} & $41.4\pm10.6$ & $31.7\pm10.1$ \\
           & Long & $69.1\pm8.8$ & {$57.1\pm9.5$} & $32.7\pm8.9$ & $38.2\pm9.2$ \\
    \midrule
    Compl. & Short & $44.7\pm8.9$ & $30.9\pm8.2$ & {$42.3\pm8.7$} & $34.1\pm8.5$ \\
    Java   & Med & $54.8\pm10.8$ & $36.6\pm10.4$ & {$47.6\pm10.8$} & $31.7\pm9.9$ \\
           & Long & $50.6\pm11.7$ & $36.9\pm11.1$ & {$23.2\pm9.6$} & $27.4\pm10.3$ \\
    \bottomrule
  \end{tabular}
  }
  \caption{Pass rate stratified by distance to the referenced object for completion and infill tasks categorized by terciles. For Python, distance boundaries for short, medium, and long are ($10$, $30$); for Java, they are ($30$, $100$). *C$3.7$ Sonnet refers to Claude $3.7$ Sonnet with extended thinking; C$3.7$ Sonnet without asterisk refers to the version without extended thinking; R1-Qwen $14$B refers to R1-Distill-Qwen-$14$B.}
  \label{tab:Length_Dep_results_combined}
\end{table}
\begin{table}[t]
  \centering
  \captionsetup{skip=5pt}
  \centerline{
  \begin{tabular}{cccccc}
    \toprule
    & Distance & O3-mini & C$3.7$ Sonnet* & DeepSeek-R1 & Qwen$2.5$Coder\\
    \midrule
    Infill & Short & {$86.0\pm4.4$} & $83.5\pm4.7$ & $75.8\pm5.5$ & $73.3\pm5.6$ \\
    Python & Long & {$78.3\pm6.9$} & $76.8\pm7.1$ & $68.9\pm7.7$ & $64.5\pm8.0$ \\
    \midrule
    Compl. & Short & $67.7\pm9.0$ & $67.7\pm9.0$ & {$68.6\pm8.9$} & $63.8\pm9.2$ \\
    Python & Long & $64.7\pm9.3$ & {$66.7\pm9.1$} & $60.8\pm9.5$ & $64.7\pm9.3$ \\
    \midrule                                
    Infill & Short & $88.4\pm4.4$ & {$89.3\pm4.2$} & $78.2\pm5.6$ & $68.4\pm6.3$ \\
    Java & Long & {$88.8\pm7.2$} & $83.3\pm8.7$ & $75.1\pm10.0$ & $47.2\pm11.7$ \\
    \midrule
    Compl. & Short & $64.8\pm6.8$ & {$68.4\pm6.6$} & $60.1\pm6.9$ & $59.6\pm6.9$ \\
    Java & Long & $47.1\pm10.6$ & {$70.6\pm9.7$} & $61.1\pm10.4$ & $48.3\pm10.5$ \\
    \midrule
    & & GPT-4o & C$3$ Opus & C$3.7$ Sonnet & C$3.5$ Haiku \\
    \midrule
    Infill & Short & $78.8\pm5.2$ & $75.5\pm5.5$ & {$75.9\pm5.5$} & $75.0\pm5.5$ \\
    Python & Long & {$77.6\pm7.0$} & $74.7\pm7.3$ & {$71.8\pm7.6$} & $64.5\pm8.0$ \\
    \midrule
    Compl. & Short & {$72.4\pm8.6$} & $66.6\pm9.0$ & $57.2\pm9.4$ & $59.0\pm9.3$ \\
    Python & Long & $66.6\pm9.2$ & {$67.6\pm9.2$} & $57.0\pm9.5$ & $60.8\pm9.5$ \\
    \midrule
    Infill & Short & $83.6\pm5.1$ & $73.3\pm6.1$ & {$75.2\pm5.9$} & $69.9\pm6.2$ \\
    Java & Long & $61.0\pm11.1$ & $49.9\pm11.6$ & {$61.1\pm11.4$} & $59.7\pm11.2$ \\
    \midrule
    Compl. & Short & $65.2\pm6.7$ & {$68.4\pm6.6$} & $71.6\pm6.4$ & $64.2\pm6.8$ \\
    Java & Long & $30.6\pm9.8$ & {$70.6\pm9.7$} & $66.0\pm10.1$ & $53.0\pm10.5$ \\
    \midrule
    & & Llama$3.3$ $70$B & Qwen-QwQ-$32$B & R1-Qwen $14$B & Llama$3.1$ $8$B \\
    \midrule
    Infill & Short & {$70.4\pm5.9$} & $61.1\pm6.2$ & $52.2\pm6.4$ & $45.3\pm6.5$ \\
    Python & Long & {$38.4\pm8.2$} & $39.1\pm8.2$ & $35.5\pm8.0$ & $40.6\pm8.2$ \\
    \midrule
    Compl. & Short & $55.2\pm9.7$ & {$48.6\pm8.9$} & $39.0\pm9.4$ & $40.0\pm9.3$ \\
    Python & Long & $52.0\pm9.8$ & {$45.1\pm9.5$} & $37.2\pm9.4$ & $37.4\pm9.3$ \\
    \midrule
    Infill & Short & {$69.9\pm6.3$} & $56.3\pm6.8$ & $42.7\pm6.8$ & $42.7\pm6.8$ \\
    Java & Long & {$52.8\pm11.4$} & $38.9\pm11.3$ & $26.3\pm10.2$ & $20.9\pm9.4$ \\
    \midrule
    Compl. & Short & {$58.0\pm6.9$} & $40.4\pm7.0$ & $49.2\pm7.1$ & $37.3\pm6.7$ \\
    Java & Long & {$29.4\pm9.7$} & $19.9\pm8.6$ & $15.2\pm7.7$ & $18.8\pm8.4$ \\
    \bottomrule
  \end{tabular}
  }
  \caption{Pass rate stratified by distance to nearest comment for completion and infill tasks. C$3.7$ Sonnet* refers to Claude $3.7$ Sonnet with extended thinking; C$3.7$ Sonnet without asterisk refers to the version without extended thinking; R1-Qwen $14$B refers to R1-Distill-Qwen-$14$B.}
  \label{tab:comments_combined}
\end{table}
\begin{table*}[!ht]
\setlength{\tabcolsep}{3.5pt}
\centering
% \label{tab:grouped_horizon_reason_cat1}
\begin{tabular}{llcccc}
\toprule

Task & Model & Python Infill & Python Compl. & Java Infill & Java Compl. \\
\midrule
Local & O3-mini    & $83.0 \pm 3.8$ & $66.0 \pm 6.5$ & $87.9 \pm 4.2$ & $56.0 \pm 6.6$ \\
Variable & C$3.7$ Sonnet* & $80.8 \pm 4.0$ & $67.0 \pm 6.4$ & $87.5 \pm 4.3$ & $65.6 \pm 6.3$ \\
& DeepSeek-R1 & $73.6 \pm 4.5$ & $64.5 \pm 6.5$ & $78.9 \pm 5.2$ & $57.8 \pm 6.7$ \\
& Qwen$2.5$Coder & $69.8 \pm 4.8$ & $64.1 \pm 6.5$ & $61.6 \pm 6.3$ & $57.4 \pm 6.5$ \\

\midrule
Global & O3-mini & \multicolumn{2}{c}{N/A} & $85.3 \pm 5.5$ & $66.3 \pm 6.9$ \\
Variable & C$3.7$ Sonnet* & \multicolumn{2}{c}{N/A} & $83.5 \pm 5.7$ & $75.6 \pm 6.2$ \\
& DeepSeek-R1 & \multicolumn{2}{c}{N/A} & $74.6 \pm 6.8$ & $62.0 \pm 7.0$ \\
& Qwen$2.5$Coder & \multicolumn{2}{c}{N/A} & $65.0 \pm 7.5$ & $53.8 \pm 7.2$ \\

\midrule
Function & O3-mini & \multicolumn{2}{c}{$73.9 \pm 12.7$} & $89.1 \pm 5.8$ & $52.2 \pm 10.4$ \\
& C$3.7$ Sonnet* & \multicolumn{2}{c}{$74.0 \pm 12.5$} & $89.1 \pm 5.8$ & $70.5 \pm 9.5$ \\
& DeepSeek-R1 & \multicolumn{2}{c}{$60.8 \pm 14.2$} & $80.9 \pm 7.3$ & $62.5 \pm 10.1$ \\
& Qwen$2.5$Coder & \multicolumn{2}{c}{$60.8 \pm 14.2$} & $61.0 \pm 9.1$ & $57.9 \pm 10.2$ \\

% \midrule
% Function & O3-mini & $81.1 \pm 12.5$ & $44.5 \pm 32.2$ & $89.1 \pm 5.8$ & $52.2 \pm 10.4$ \\
% & Claude 3.7 Sonnet(ET) & $81.1 \pm 12.8$ & $44.2 \pm 32.7$ & $89.1 \pm 5.8$ & $70.5 \pm 9.5$ \\
% & DeepSeek-R1 & $67.5 \pm 15.2$ & $33.6 \pm 31.0$ & $80.9 \pm 7.3$ & $62.5 \pm 10.1$ \\
% & Qwen2.5Coder-32B & $67.5 \pm 15.2$ & $33.3 \pm 31.0$ & $61.0 \pm 9.1$ & $57.9 \pm 10.2$ \\

\midrule
Class & O3-mini & \multicolumn{2}{c}{$47.0 \pm 16.9$} & $83.4 \pm 10.0$ & $65.7 \pm 16.6$ \\
& C$3.7$ Sonnet* & \multicolumn{2}{c}{$55.8 \pm 16.7$} & $81.5 \pm 10.3$ & $59.5 \pm 16.8$ \\
& DeepSeek-R1 & \multicolumn{2}{c}{$53.0 \pm 16.8$} & $77.7 \pm 11.1$ & $65.6 \pm 16.6$ \\
& Qwen$2.5$Coder & \multicolumn{2}{c}{$41.3 \pm 16.7$} & $61.1 \pm 12.9$ & $43.7 \pm 17.0$ \\
% \midrule
% Class & O3-mini & $73.2 \pm 22.3$ & $26.4 \pm 19.8$ & $83.4 \pm 10.0$ & $65.7 \pm 16.6$ \\
% & Claude 3.7 Sonnet(ET) & $73.2 \pm 22.4$ & $42.1 \pm 22.0$ & $81.5 \pm 10.3$ & $59.5 \pm 16.8$ \\
% & DeepSeek-R1 & $86.7 \pm 17.1$ & $26.4 \pm 19.9$ & $77.7 \pm 11.1$ & $65.6 \pm 16.6$ \\
% & Qwen2.5Coder-32B & $59.9 \pm 25.1$ & $26.5 \pm 19.8$ & $61.1 \pm 12.9$ & $43.7 \pm 17.0$ \\

\midrule
Library & O3-mini & $77.2 \pm 5.8$ & $61.1 \pm 8.7$ & \multicolumn{2}{c}{N/A} \\
& C$3.7$ Sonnet* & $71.8 \pm 6.2$ & $65.8 \pm 8.5$ & \multicolumn{2}{c}{N/A} \\
& DeepSeek-R1 & $69.3 \pm 6.4$ & $57.7 \pm 8.8$ & \multicolumn{2}{c}{N/A} \\
& Qwen$2.5$Coder & $60.3 \pm 6.8$ & $59.3 \pm 8.6$ & \multicolumn{2}{c}{N/A} \\

\midrule
If-Else & O3-mini & \multicolumn{2}{c}{$51.6 \pm 17.3$} & $78.6 \pm 10.9$ & $54.5 \pm 17.0$ \\
Condition & C$3.7$ Sonnet & \multicolumn{2}{c}{$48.5 \pm 16.9$} & $87.5 \pm 8.7$ & $63.6 \pm 16.8$ \\
& DeepSeek-R1 & \multicolumn{2}{c}{$45.4 \pm 17.2$} & $66.1 \pm 12.4$ & $48.6 \pm 16.6$ \\
& Qwen$2.5$Coder & \multicolumn{2}{c}{$30.4 \pm 15.7$} & $64.3 \pm 12.6$ & $48.4 \pm 16.9$ \\

% \midrule
% If-Else & O3-mini & $57.9 \pm 22.2$ & $42.8 \pm 25.6$ & $78.6 \pm 10.9$ & $54.5 \pm 17.0$ \\
% Condition & Claude 3.7 Sonnet(ET) & $58.0 \pm 22.3$ & $35.5 \pm 25.1$ & $87.5 \pm 8.7$ & $63.6 \pm 16.8$ \\
% & DeepSeek-R1 & $47.4 \pm 22.5$ & $50.0 \pm 26.5$ & $66.1 \pm 12.4$ & $48.6 \pm 16.6$ \\
% & Qwen2.5Coder-32B & $26.2 \pm 20.0$ & $35.7 \pm 25.1$ & $64.3 \pm 12.6$ & $48.4 \pm 16.9$ \\

\midrule
If-Else & O3-mini & $80.3 \pm 10.0$ & $71.3 \pm 10.4$ & $91.2 \pm 6.3$ & $60.0 \pm 8.7$ \\
Body & C$3.7$ Sonnet* & $78.7 \pm 10.3$ & $71.3 \pm 10.3$ & $88.6 \pm 7.0$ & $65.8 \pm 8.6$ \\
& DeepSeek-R1 & $68.9 \pm 11.6$ & $69.9 \pm 10.5$ & $82.3 \pm 8.5$ & $61.7 \pm 8.6$ \\
& Qwen$2.5$Coder & $65.4 \pm 11.8$ & $61.6 \pm 11.1$ & $70.9 \pm 10.0$ & $57.5 \pm 8.8$ \\

\midrule
Loop & O3-mini & $83.3 \pm 8.6$ & $71.5 \pm 12.6$ & $90.2 \pm 5.8$ & $57.8 \pm 9.6$ \\
Body & C$3.7$ Sonnet* & $80.6 \pm 9.1$ & $67.3 \pm 13.0$ & $89.2 \pm 6.0$ & $58.8 \pm 9.5$ \\
& DeepSeek-R1 & $69.5 \pm 10.6$ & $67.3 \pm 13.1$ & $87.3 \pm 6.5$ & $57.9 \pm 9.6$ \\
& Qwen$2.5$Coder & $63.9 \pm 11.3$ & $59.1 \pm 13.9$ & $62.7 \pm 9.4$ & $56.8 \pm 9.6$ \\

\bottomrule
\end{tabular}
\caption{Pass rates stratified by task type. Note that some results marked N/A are not presented due to insufficient data or overly-large confidence intervals. C$3.7$ Sonnet* refers to Claude $3.7$ Sonnet with extended thinking with 16K token budget, and Qwen$2.5$Coder-$32$B is abbreviated as Qwen$2.5$Coder.}
\label{tab:by-construct1}
\end{table*}

\begin{table*}[!ht]
\setlength{\tabcolsep}{3.5pt}
\centering
% \label{tab:grouped_horizon_reason_cat2}
\begin{tabular}{llcccc}
\toprule
Task & Model & Python Infill & Python Compl. & Java Infill & Java Compl. \\
\midrule
% Local/Variable Block:
Local & GPT-4o    & $78.2 \pm 4.2$ & $69.4 \pm 6.3$ & $77.2 \pm 5.3$ & $57.8 \pm 6.5$ \\
Variable & C3 Opus & $75.5 \pm 4.4$ & $66.9 \pm 6.4$ & $65.9 \pm 6.1$ & $66.5 \pm 6.3$ \\
& C$3.7$ Sonnet & $74.1 \pm 4.4$ & $56.8 \pm 6.8$ & $71.1 \pm 5.9$ & $68.4 \pm 6.2$ \\
& C$3.5$ Haiku & $70.9 \pm 4.6$ & $60.2 \pm 6.7$ & $65.5 \pm 6.1$ & $55.5 \pm 6.6$ \\
\midrule
% Global/Variable Block (unchanged):
Global & GPT-4o & \multicolumn{2}{c}{N/A} & $79.0 \pm 6.4$ & $55.0\pm7.2$ \\
Variable & C3 Opus & \multicolumn{2}{c}{N/A} & $66.3 \pm 7.4$ & $69.5\pm6.7$ \\
& C$3.7$ Sonnet & \multicolumn{2}{c}{N/A} & $73.9 \pm 6.8$ & $71.2\pm6.5$ \\
& C$3.5$ Haiku & \multicolumn{2}{c}{N/A} & $67.6 \pm 7.3$ & $67.9\pm6.7$ \\
\midrule
% Function Block:
Function & GPT-4o & \multicolumn{2}{c}{$69.6 \pm 13.3$} & $73.6 \pm 8.2$ & $47.7\pm10.4$ \\
& C3 Opus & \multicolumn{2}{c}{$65.2 \pm 13.8$} & $62.8 \pm 9.0$ & $64.4\pm10.0$ \\
& C$3.7$ Sonnet & \multicolumn{2}{c}{$65.2 \pm 13.8$} & $73.6 \pm 8.2$ & $67.7\pm9.8$ \\
& C$3.5$ Haiku & \multicolumn{2}{c}{$65.2 \pm 13.8$} & $61.9 \pm 9.0$ & $48.9\pm10.5$ \\
\midrule
% % Function Block:
% Function & GPT-4o & $72.9 \pm 14.4$ & $55.7 \pm 32.4$ & $73.6 \pm 8.2$ & $47.7\pm10.4$ \\
% & C3 Opus & $72.9 \pm 14.4$ & $33.4 \pm 30.6$ & $62.8 \pm 9.0$ & $64.4\pm10.0$ \\
% & C3.7 Sonnet & $70.4 \pm 14.8$ & $44.2 \pm 32.2$ & $73.6 \pm 8.2$ & $67.7\pm9.8$ \\
% & C3.5 Haiku & $70.4 \pm 14.8$ & $44.3 \pm 32.4$ & $61.9 \pm 9.0$ & $48.9\pm10.5$ \\
% \midrule
% Class Block:
Class & GPT-4o & \multicolumn{2}{c}{$55.9 \pm 16.7$} & $79.7\pm10.7$ & $53.0\pm17.3$ \\
& C3 Opus & \multicolumn{2}{c}{$58.8 \pm 16.5$} & $61.0\pm13.0$ & $49.9\pm17.4$ \\
& C$3.7$ Sonnet & \multicolumn{2}{c}{$41.2 \pm 16.5$} & $66.6\pm12.7$ & $62.5\pm16.5$ \\
& C$3.5$ Haiku & \multicolumn{2}{c}{$41.2 \pm 16.5$} & $49.9\pm13.5$ & $46.9\pm17.2$ \\
\midrule
% % Class Block:
% Class & GPT-4o & $86.7 \pm 17.3$ & $31.6 \pm 20.8$ & $79.7\pm10.7$ & $53.0\pm17.3$ \\
% & C3 Opus & $73.2 \pm 22.4$ & $47.4 \pm 22.4$ & $61.0\pm13.0$ & $49.9\pm17.4$ \\
% & C3.7 Sonnet & $66.3 \pm 23.7$ & $21.1 \pm 18.3$ & $66.6\pm12.7$ & $62.5\pm16.5$ \\
% & C3.5 Haiku & $60.0 \pm 25.0$ & $26.2 \pm 19.8$ & $49.9\pm13.5$ & $46.9\pm17.2$ \\
% \midrule
% Library Block (Java columns N/A):
Library & GPT-4o & $71.3 \pm 6.2$ & $65.0 \pm 8.4$ & \multicolumn{2}{c}{N/A} \\
& C3 Opus & $67.8 \pm 6.4$ & $63.4 \pm 8.5$ & \multicolumn{2}{c}{N/A} \\
& C$3.7$ Sonnet & $66.8 \pm 6.6$ & $52.9 \pm 8.7$ & \multicolumn{2}{c}{N/A} \\
& C$3.5$ Haiku & $63.4 \pm 6.6$ & $56.1 \pm 8.7$ & \multicolumn{2}{c}{N/A} \\
\midrule
% If-Else / Condition Block:
If-Else & GPT-4o & \multicolumn{2}{c}{$54.5 \pm 17.0$} & $82.1 \pm 10.0$ & $60.7 \pm 16.7$ \\
Condition & C3 Opus & \multicolumn{2}{c}{$36.4 \pm 16.4$} & $73.2 \pm 11.7$ & $63.6 \pm 16.4$ \\
& C$3.7$ Sonnet & \multicolumn{2}{c}{$51.5 \pm 17.1$} & $78.6 \pm 10.9$ & $69.8 \pm 15.7$ \\
& C$3.5$ Haiku & \multicolumn{2}{c}{$30.3 \pm 15.7$} & $76.7 \pm 11.1$ & $57.5 \pm 17.0$ \\
\midrule
% % If-Else / Condition Block:
% If-Else & GPT-4o & $68.3 \pm 21.0$ & $35.9 \pm 25.4$ & $82.1 \pm 10.0$ & $60.7 \pm 16.7$ \\
% Condition & C3 Opus & $31.6 \pm 20.8$ & $42.7 \pm 25.7$ & $73.2 \pm 11.7$ & $63.6 \pm 16.4$ \\
% & C3.7 Sonnet & $68.3 \pm 21.1$ & $28.4 \pm 23.8$ & $78.6 \pm 10.9$ & $69.8 \pm 15.7$ \\
% & C3.5 Haiku & $36.9 \pm 21.8$ & $21.5 \pm 21.5$ & $76.7 \pm 11.1$ & $57.5 \pm 17.0$ \\
% \midrule
% If-Else Body Block:
If-Else & GPT-4o & $81.9 \pm 9.7$ & $71.2 \pm 10.5$ & $79.7 \pm 8.9$ & $65.0 \pm 8.6$ \\
Body & C3 Opus & $62.3 \pm 12.4$ & $64.3 \pm 10.9$ & $72.2 \pm 10.0$ & $65.0 \pm 8.5$ \\
& C$3.7$ Sonnet & $78.7 \pm 10.3$ & $58.9 \pm 11.2$ & $74.7 \pm 9.5$ & $72.5 \pm 7.9$ \\
& C$3.5$ Haiku & $60.7 \pm 12.3$ & $53.4 \pm 11.5$ & $74.7 \pm 9.7$ & $64.2 \pm 8.4$ \\
\midrule
% Loop Body Block:
Loop & GPT-4o & $77.8 \pm 9.7$ & $71.5 \pm 12.8$ & $80.4 \pm 7.7$ & $61.8 \pm 9.4$ \\
Body & C3 Opus & $66.7 \pm 10.9$ & $67.3 \pm 13.0$ & $63.7 \pm 9.3$ & $61.7 \pm 9.4$ \\
& C$3.7$ Sonnet & $76.4 \pm 9.8$ & $67.3 \pm 13.1$ & $75.5 \pm 8.3$ & $63.7 \pm 9.4$ \\
& C$3.5$ Haiku & $73.6 \pm 10.2$ & $59.1 \pm 13.9$ & $68.6 \pm 9.0$ & $60.8 \pm 9.5$ \\
\bottomrule
\end{tabular}
\caption{Pass rates stratified by task type. For Java, both Infill and Completion are not applicable for the Library task. C$3.7$ Sonnet refers to Claude $3.7$ Sonnet without extended thinking.}
\label{tab:by-construct2}
\end{table*}

\begin{table*}[!ht]
\setlength{\tabcolsep}{3.5pt}
\centering
% \label{tab:grouped_horizon_reason_cat3}
\begin{tabular}{llcccc}
\toprule
Task & Model & Python Infill & Python Compl. & Java Infill & Java Compl. \\
\midrule
% --- Local/Variable Block ---
Local & Llama$3.3$ $70$B    & $41.8 \pm 5.0$  & $49.1 \pm 6.9$  & $42.7 \pm 6.4$ & $43.6 \pm 6.6$ \\
Variable & Qwen-QwQ-$32$B & $52.8 \pm 5.1$  & $46.6 \pm 6.7$  & $51.7 \pm 6.4$ & $34.9 \pm 6.4$ \\
& R1-Qwen $14$B & $46.4 \pm 5.1$  & $38.3 \pm 6.7$  & $38.4 \pm 6.3$ & $39.4 \pm 6.6$ \\
& Llama$3.1$ $8$B & $43.1 \pm 5.0$  & $38.9 \pm 6.7$  & $34.9 \pm 6.2$ & $30.8 \pm 6.1$ \\
\midrule
% --- Global/Variable Block (Python: N/A) ---
Global & Llama$3.3$ $70$B & \multicolumn{2}{c}{N/A} & $37.6 \pm 7.6$ & $48.3 \pm 7.1$ \\
Variable & Qwen-QwQ-$32$B & \multicolumn{2}{c}{N/A} & $50.4 \pm 7.8$ & $36.9 \pm 7.0$ \\
& R1-Qwen $14$B & \multicolumn{2}{c}{N/A} & $36.9 \pm 7.6$ & $43.0 \pm 7.2$ \\
& Llama$3.1$ $8$B & \multicolumn{2}{c}{N/A} & $33.8 \pm 7.4$ & $32.1 \pm 6.8$ \\
\midrule
% --- Function Block ---
Function & Llama$3.3$ $70$B & \multicolumn{2}{c}{$58.7 \pm 14.2$} & $53.6 \pm 9.4$  & $42.1 \pm 10.4$ \\
& Qwen-QwQ-$32$B & \multicolumn{2}{c}{$43.5 \pm 14.3$} & $50.9 \pm 9.3$  & $31.8 \pm 9.7$ \\
& R1-Qwen $14$B & \multicolumn{2}{c}{$41.3 \pm 14.2$} & $39.0 \pm 9.1$  & $30.6 \pm 9.6$ \\
& Llama$3.1$ $8$B & \multicolumn{2}{c}{$39.1 \pm 14.1$} & $31.0 \pm 8.7$  & $26.2 \pm 9.2$ \\
\midrule
% % --- Function Block ---
% Function & Llama3.3 70B & $56.8 \pm 16.0$  & $22.2 \pm 27.2$ & $53.6 \pm 9.4$  & $42.1 \pm 10.4$ \\
% & Qwen QwQ 32B & $48.7 \pm 16.0$  & $22.3 \pm 27.1$ & $50.9 \pm 9.3$  & $31.8 \pm 9.7$ \\
% & DS Qwen 14B & $48.8 \pm 15.9$  & $11.0 \pm 20.5$ & $39.0 \pm 9.1$  & $30.6 \pm 9.6$ \\
% & Llama3.1 8B & $46.1 \pm 16.0$  & $11.1 \pm 20.5$ & $31.0 \pm 8.7$  & $26.2 \pm 9.2$ \\
% \midrule
% --- Class Block ---
Class & Llama$3.3$ $70$B & \multicolumn{2}{c}{$11.8 \pm 10.8$} & $59.4 \pm 13.1$ & $43.7 \pm 17.2$ \\
& Qwen-QwQ-$32$B & \multicolumn{2}{c}{$29.4 \pm 15.3$} & $57.5 \pm 13.3$ & $43.8 \pm 17.1$ \\
& R1-Qwen $14$B & \multicolumn{2}{c}{$14.7 \pm 11.9$} & $24.0 \pm 11.4$ & $31.3 \pm 16.0$ \\
& Llama$3.1$ $8$B & \multicolumn{2}{c}{$23.5 \pm 14.3$} & $38.9 \pm 13.0$ & $28.0 \pm 15.7$ \\
\midrule
% % --- Class Block ---
% Class & Llama3.3 70B & $6.8 \pm 12.7$   & $15.8 \pm 16.5$ & $59.4 \pm 13.1$ & $43.7 \pm 17.2$ \\
% & Qwen QwQ 32B & $40.1 \pm 24.7$  & $21.1 \pm 18.4$ & $57.5 \pm 13.3$ & $43.8 \pm 17.1$ \\
% & DS Qwen 14B & $13.2 \pm 17.0$  & $15.8 \pm 16.5$ & $24.0 \pm 11.4$ & $31.3 \pm 16.0$ \\
% & Llama3.1 8B & $40.2 \pm 24.7$  & $10.6 \pm 13.9$ & $38.9 \pm 13.0$ & $28.0 \pm 15.7$ \\
% \midrule
% --- Library Block (Java: N/A) ---
Library & Llama$3.3$ $70$B & $35.6 \pm 6.5$   & $38.2 \pm 8.7$  & \multicolumn{2}{c}{N/A} \\
& Qwen-QwQ-$32$B & $47.5 \pm 6.8$   & $42.3 \pm 8.8$  & \multicolumn{2}{c}{N/A} \\
& R1-Qwen $14$B & $36.7 \pm 6.7$   & $29.3 \pm 8.0$  & \multicolumn{2}{c}{N/A} \\
& Llama3.1 8B & $38.1 \pm 6.7$   & $29.3 \pm 8.0$  & \multicolumn{2}{c}{N/A} \\
\midrule
% --- If-Else / Condition Block ---
If-Else & Llama$3.3$ $70$B & \multicolumn{2}{c}{$30.3 \pm 15.7$} & $50.0 \pm 13.0$ & $45.6 \pm 17.0$ \\
Condition & Qwen-QwQ-$32$B & \multicolumn{2}{c}{$18.2 \pm 13.2$} & $39.3 \pm 12.7$ & $30.2 \pm 15.9$ \\
& R1-Qwen $14$B & \multicolumn{2}{c}{$27.3 \pm 15.2$} & $37.4 \pm 12.7$ & $42.4 \pm 16.7$ \\
& Llama$3.1$ $8$B & \multicolumn{2}{c}{$12.1 \pm 11.1$} & $42.9 \pm 13.0$ & $33.2 \pm 15.9$ \\
\midrule
% % --- If-Else / Condition Block ---
% If-Else & Llama3.3 70B & $20.9 \pm 18.4$  & $14.3 \pm 18.3$ & $50.0 \pm 13.0$ & $45.6 \pm 17.0$ \\
% Condition & Qwen QwQ 32B & $26.4 \pm 19.7$  & $7.2 \pm 13.5$  & $39.3 \pm 12.7$ & $30.2 \pm 15.9$ \\
% & DS Qwen 14B & $31.6 \pm 20.7$  & $21.6 \pm 21.5$ & $37.4 \pm 12.7$ & $42.4 \pm 16.7$ \\
% & Llama3.1 8B & $5.2 \pm 9.9$    & $21.5 \pm 21.6$ & $42.9 \pm 13.0$ & $33.2 \pm 15.9$ \\
% \midrule
% --- If-Else Body Block ---
If-Else & Llama$3.3$ $70$B & $32.8 \pm 12.0$  & $52.0 \pm 11.5$ & $59.5 \pm 10.8$ & $48.4 \pm 8.8$ \\
Body &Qwen-QwQ-$32$B & $39.4 \pm 12.4$  & $43.8 \pm 11.4$ & $55.7 \pm 10.9$ & $42.4 \pm 8.8$ \\
& R1-Qwen $14$B & $41.0 \pm 12.4$  & $37.0 \pm 11.1$ & $44.2 \pm 11.1$ & $47.5 \pm 9.0$ \\
& Llama$3.1$ $8$B & $32.9 \pm 11.8$  & $43.8 \pm 11.4$ & $38.0 \pm 10.7$ & $40.0 \pm 8.8$ \\
\midrule
% --- Loop Body Block ---
Loop & Llama$3.3$ $70$B & $50.0 \pm 11.7$  & $38.8 \pm 13.6$ & $51.9 \pm 9.6$  & $46.0 \pm 9.6$ \\
Body & Qwen-QwQ-$32$B & $51.4 \pm 11.6$  & $36.7 \pm 13.4$ & $55.9 \pm 9.6$  & $39.2 \pm 9.5$ \\
& R1-Qwen $14$B & $37.5 \pm 11.3$  & $28.6 \pm 12.8$ & $45.1 \pm 9.6$  & $42.1 \pm 9.6$ \\
& Llama$3.1$ $8$B & $25.0 \pm 10.1$  & $28.5 \pm 12.8$ & $43.1 \pm 9.5$  & $34.3 \pm 9.2$ \\
\bottomrule
\end{tabular}
\caption{Pass rates stratified by task type. For Java, both Infill and Completion are not applicable for the Library task. R1-Qwen $14$B refers to R1-Distill-Qwen-$14$B.}
\label{tab:by-construct3}
\end{table*}

\end{document}